\documentclass{article}

\usepackage{microtype}
\usepackage{graphicx}
\usepackage{booktabs} 
\usepackage{multirow}
\usepackage{amsmath}
\usepackage{amsfonts}
\usepackage{amsthm}
\usepackage{array}
\usepackage{subcaption}
\usepackage{hyperref}       
\usepackage{url}            
\usepackage{bm}
\usepackage{textcomp}
\usepackage{caption}
\usepackage{rotating}

\usepackage[accepted]{icml2019}

\def \xb {\mathbf{x}}
\def \xa {\mathbf{x}_{adv}}
\def \zb {\mathbf{z}}
\def \pb {\mathbf p}

\def \cU {\mathcal{U}}

\def \btheta {{\bm \theta}}

\def \gb {\mathbf{g}}

\newcommand{\argmax}{\mathop{\mathrm{arg\,max}}}

\icmltitlerunning{Understanding Adversarial Attacks on Deep Learning Based Medical Image Analysis Systems}

\begin{document}

\twocolumn[
\icmltitle{Understanding Adversarial Attacks on Deep Learning Based \\ Medical Image Analysis Systems}



\icmlsetsymbol{equal}{*,}
\icmlsetsymbol{cor}{**,}
\icmlsetsymbol{buaa}{a,}
\icmlsetsymbol{melb}{b}
\icmlsetsymbol{bigdata}{c}
\icmlsetsymbol{nii}{d}
\icmlsetsymbol{shjt}{e}
\icmlsetsymbol{cixi}{f}

\begin{icmlauthorlist}
\icmlauthor{Xingjun Ma}{equal,melb}
\icmlauthor{Yuhao Niu}{equal,buaa,bigdata}
\icmlauthor{Lin Gu}{nii}
\icmlauthor{Yisen Wang}{shjt}
\icmlauthor{Yitian Zhao}{cixi}
\icmlauthor{James Bailey}{melb}
\icmlauthor{Feng Lu}{cor,buaa,bigdata}
\end{icmlauthorlist}

\icmlcorrespondingauthor{Feng Lu}{lufeng@buaa.edu.cn}

\begin{center}
    \small
    \textsuperscript{a}State Key Laboratory of VR Technology and Systems, School of CSE, Beihang University, Beijing, China. \\
    \textsuperscript{b}School of Computing and Information Systems, The University of Melbourne, Parkville, VIC 3010, Australia. \\
    \textsuperscript{c}Beijing Advanced Innovation Center for Big Data-Based Precision Medicine, Beihang University, Beijing, China. \\
    \textsuperscript{d}National Institute of Informatics, Tokyo 101-8430, Japan. \\
    \textsuperscript{e}Department of Computer Science and Engineering, Shanghai Jiao Tong University, Shanghai, China. \\
    \textsuperscript{f}Cixi Instuitue of Biomedical Engineering, Ningbo Institute of Industrial Technology, Chinese Academy of Sciences, Ningbo, China.
\end{center}
\icmlkeywords{Adversarial Attack, Adversarial Example Detection, Medical Image Analysis, Deep Learning}

\vskip 0.3in
]



\printAffiliationsAndNotice{\icmlEqualContribution} 

\begin{abstract}
Deep neural networks (DNNs) have become popular for medical image analysis tasks like cancer diagnosis and lesion detection. 
However, a recent study demonstrates that medical deep learning systems can be compromised by carefully-engineered adversarial examples/attacks with small imperceptible perturbations.
This raises safety concerns about the deployment of these systems in clinical settings. In this paper, we provide a deeper understanding of adversarial examples in the context of medical images. 
We find that medical DNN models can be more vulnerable to adversarial attacks compared to models for natural images, according to two different viewpoints. 
Surprisingly, we also find that medical adversarial attacks can be easily detected, i.e., simple detectors can achieve over 98\% detection AUC against state-of-the-art attacks, due to fundamental feature differences compared to normal examples. 
We believe these findings may be a useful basis to approach the design of more explainable and secure medical deep learning systems.
\end{abstract}

\section{Introduction}
Deep neural networks (DNNs) are powerful models that have been widely used to achieve near human-level performance on a variety of natural image analysis tasks such as image classification \cite{he2016deep}, object detection \cite{Wang19IGRSL}, image retrieval \cite{Bai_PR2018} and 3D analysis \cite{LuCSS18}. Driven by their current success on natural images (eg. images captured from natural scenes such as CIFAR-10 and ImageNet), DNNs have become a popular tool for medical image processing tasks, such as cancer diagnosis \cite{esteva2017dermatologist}, diabetic retinopathy detection \cite{kaggle2015diabetic} and organ/landmark localization \cite{roth2015deeporgan}.
Despite their superior performance, 
recent studies have found that state-of-the-art DNNs are vulnerable to carefully crafted adversarial examples (or attacks), \textit{i.e.}, slightly perturbed input instances can fool DNNs into making incorrect predictions with high confidence \cite{szegedy2013intriguing,goodfellow2014explaining}.
This has raised safety concerns about the deployment of deep learning models in safety-critical applications such as autonomous driving \cite{evtimov2017robust}, action analysis \cite{ChengLZ18} and medical diagnosis \cite{finlayson2019adversarial}.

\begin{figure*}[!t]
    \centering
    \includegraphics[width=0.6\textwidth]{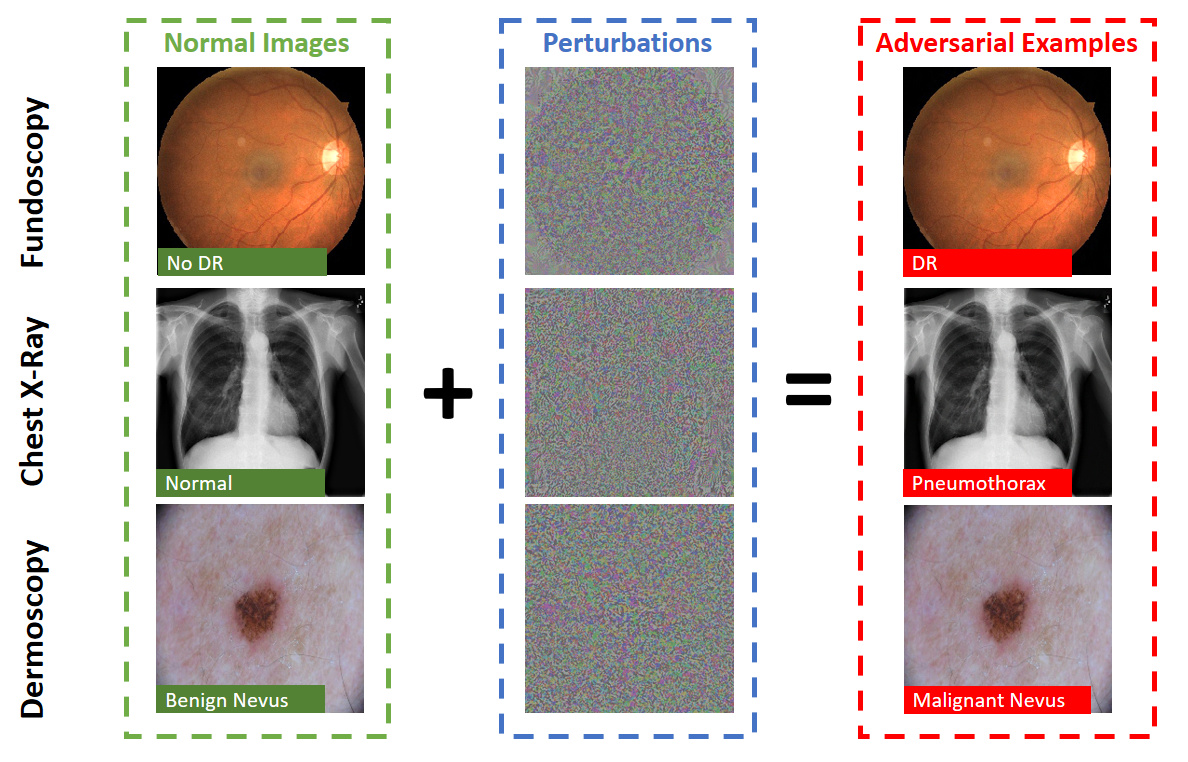}
    \caption{Examples of adversarial attacks crafted by the Projected Gradient Descent (PGD) to fool DNNs trained on medical image datasets Fundoscopy \cite{kaggle2015diabetic} (first row, DR=diabetic retinopathy), Chest X-Ray \cite{wang2017chestxray} (second row) and Dermoscopy \cite{isic_archive} (third row). \emph{Left}: normal images, \emph{Middle}: adversarial perturbations, \emph{Right}: adversarial images. The left bottom tag is the predicted class, and green/red indicates correct/wrong predictions.}
    \label{fig:examples}
\end{figure*}

While existing works on adversarial machine learning research have mostly focused on natural images, a full understanding of adversarial attacks in the medical image domain is still open. Medical images can have domain-specific characteristics that are quite different from natural images, for example, unique biological textures.
A recent work has confirmed that medical deep learning systems can also be compromised by adversarial attacks \cite{finlayson2019adversarial}. As shown in Figure \ref{fig:examples}, across three medical image datasets Fundoscopy \cite{kaggle2015diabetic}, Chest X-Ray \cite{wang2017chestxray} and Dermoscopy \cite{isic_archive}, diagnosis results can be arbitrarily manipulated by adversarial attacks.
Such a vulnerability has also been discussed in 3D volumetric medical image segmentation \cite{li2019volumetric}.
Considering the vast sums of money which underpin the healthcare economy, this inevitably creates risks whereby potential attackers may seek to profit from manipulation against the healthcare system. For example, an attacker might manipulate their examination reports to commit insurance fraud or a false claim of medical reimbursement \cite{paschali2018generalizability}. On the other hand, an attacker might seek to cause disruption by imperceptibly manipulating an image to cause a misdiagnosis of disease. This could have severe impact for the decisions made about a patient. 
To make it worse, since the DNN works in a black-box way \cite{niu2019pathological}, this falsified decision could hardly be recognised. 
As deep learning models and medical imaging techniques become increasingly used in the process of medical diagnostics, decision support and  pharmaceutical approvals \cite{pien2005using}, secure and robust medical deep learning systems become crucial \cite{finlayson2019adversarial,paschali2018generalizability}. A first and important step is to develop a comprehensive understanding of adversarial attacks in this domain.

In this paper, we provide a comprehensive understanding of medical image adversarial attacks from the perspective of generating as well as detecting these attacks.  
Two recent works \cite{finlayson2019adversarial,paschali2018generalizability} have investigated adversarial attacks on medical images and mainly focused on testing the robustness of deep models designed for medical image analysis. In particular, the work of \cite{paschali2018generalizability}  tested whether existing medical deep learning models can be attacked by adversarial attacks.  They showed that classification accuracy drops from above $87\%$ on normal medical images to almost $0\%$ on adversarial examples. Work in \cite{paschali2018generalizability} utilized adversarial examples as a measure to evaluate the robustness of medical imaging models in classification or segmentation tasks. Their study was restricted to small perturbations and they observed a marginal but variable performance drop across different models. Despite these studies, the following question has remained open \emph{``Can adversarial attacks on medical images be crafted as easily as attacks on natural images? If not, why?''}.  Furthermore, to the best of our knowledge, no previous work has investigated the detection of medical image adversarial examples. A natural question here is to ask  \emph{``To what degree are adversarial attacks on medical images detectable?''}.   In this paper, we provide some answers to these questions by investigating both the crafting (generation) and detection of adversarial attacks on medical images.

In summary, our main contributions are:
\begin{enumerate}
    \item We find that adversarial attacks on medical images can succeed more easily than those on natural images. That is, less perturbation is required to craft a successful attack.
    
    \item We show the higher vulnerability of medical image DNNs appears to be due to several reasons: 
    1) some medical images have complex biological textures, leading to more high gradient regions that are sensitive to small adversarial perturbations; and most importantly, and 2) state-of-the-art DNNs designed for large-scale natural image processing can be overparameterized for medical imaging tasks, resulting in a sharp loss landscape and high vulnerability to adversarial attacks.
    
    \item We show that surprisingly,  medical image adversarial attacks can also be easily detected.  A simple detector trained on deep features alone can achieve over 98\% detection AUC against all tested attacks across our three datasets. To the best of our knowledge, this is the first work on the detection of adversarial attacks in the medical image domain.
    
    \item We show that the high detectability of medical image adversarial examples appears to be because adversarial attacks result in perturbations to widespread regions outside the lesion area.  This results in deep feature values for adversarial examples that are recognizably different from those of normal examples.
\end{enumerate}

Our findings of different degrees of adversarial vulnerabilities of DNNs on medical versus natural images can help develop a more comprehensive understanding on the reliability and robustness of deep learning models in different domains.
The set of reasons we identified for such a difference reveal more insights into the behavior of DNNs in the presence of different types of adversarial examples. Our analysis of medical adversarial examples provides new interpretations of the learned representations and additional explanations for the decisions made by deep learning models in the context of medical images. This is a useful starting point towards building explainable and robust deep learning systems for medical diagnosis.

The remainder of this paper is organized as follows. In section \ref{sec:related}, we briefly introduce deep learning based medical image analysis. In section \ref{sec:preliminaries}, we provide an introduction to adversarial attack and defense techniques. We conduct systematic experiments in sections 4 \& 5 to investigate and understand the behaviour of medical image adversarial attacks. Section \ref{sec:conclusion} discusses several future work and summarizes our contributions. 

\section{Background of Medical Image Analysis}\label{sec:related}

Driven by the current success of deep learning in traditional computer vision, the  field of medical imaging analysis (MIA) has also been influenced by DNN models.  One of the first contributions of DNNs was in the area of medical image classification.
This includes several highly successful applications of DNNs in medical diagnosis, such as the severity stage of diabetic retinopathy from retinal fundoscopy \cite{kaggle2015diabetic}, lung diseases from chest X-ray \cite{wang2017chestxray} or skin cancer from dermoscopic photographs \cite{isic_archive}. Another important application of DNNs in medical image analysis is the segmentation of organs or lesions. Organ segmentation aims to quantitatively measure the organs, such as vessels\cite{Gu_2015_ICCV,Liu_MICCAI19} and kidneys \cite{Wang_MIA_2019}, as a prelude to diagnosis or radiology therapy. 
Registration is another important task in medical imaging, where the objective is to spatially align medical images from different modalities or capture settings. For example, \cite{Cheng_CMBB_2015} exploited the local similarity between CT and MRI images with two types of auto-encoders.

Deep learning based medical image analysis may operate on a variety of input image sources, such as visible light images, hyperspectral light images, X-rays and nuclear magnetic resonance images, across various anatomical areas such as the brain, chest, skin and retina. Brain images have been extensively studied to diagnose Alzheimer’s disease \cite{Liu_ISBI_2015} and tumor segmentation \cite{Menze_TMI_2015}. 
Ophthalmic imaging is another important application, which mainly focuses either on color fundus imaging (CFI) or Optical coherence tomography (OCT) for eye disease diagnosis or abnormalities segmentation.  Among these applications, the deep learning based diabetic retinopathy diagnosis system was the first that was approved by the US Food and Drug Administration (FDA). 
\cite{Gulshan_JAMA_2016} achieved comparable accuracy in detecting diabetic retinopathy to seven certified ophthalmologists using an Inception network.   
There are systems that apply Convolutional Neural Networks (CNNs) to extract deep features to detect and classify nodules \cite{wang2017chestxray} in the chest from radiography and computed tomography (CT).
Digital pathology and microscopy is also a popular task due to the heavy burden on clinicians analyzing large numbers of histopathology images of tissue specimens. Specifically, this task involves segmenting high density cells and classifying the mitoses\cite{Ciresan_MICCAI_13}. The above studies rely on the images captured by specialized cameras or devices. In contrast, in the context of skin cancer, it has been shown that standard cameras can deliver excellent performance as input to DNN models \cite{esteva2017dermatologist}.
Inspired by this success, the International Skin Imaging Collaboration \cite{isic_archive} released a large dataset to support research on melanoma early detection. 

Most of these methods, especially diagnosis ones, adopt roughly the same pipeline, on a variety of images including ophthalmology \cite{kaggle2015diabetic}, radiology \cite{wang2017chestxray} and  dermatology \cite{isic_archive}. The images are input into CNNs  (typically the most advanced ones existing at the time, such as `AlexNet', `VGG', `Inception' and `ResNet' \cite{he2016deep}) to learn intermediate medical features before generating the final output. Whilst these pipelines have achieved excellent success, similar to those for standard computer vision object recognition, they have been criticized for having a lack of transparency. 
Though some preliminary attempt \cite{niu2019pathological}, has been proposed to use Koch postulates, the foundation of evidence based medicine, to explore the decision made by DNNs.  People still
find it difficult to verify the system's reasoning, which is essential for clinical applications which require high levels of trust. It is easy to see that such trust may be further eroded  by the existence of adversarial examples, whereby an imperceptible modification may result in costly and sometimes irreparable damage. 
We next discuss methods for adversarial attack and detection.

\section{Preliminaries}\label{sec:preliminaries}
In this paper, we focus on medical image classification tasks using DNNs. For a $K$-class ($K \geq 2$) classification problem, given a dataset $\{(\xb_i,y_i)\}_{i=1,\dots,N}$ with $\xb_i\in \mathbb{R}^d$ as a normal example and $y_i\in\{1,\dots,K\}$ as its associated label, a DNN classifier $h$ with parameter $\btheta$ predicts the class of an input example $\xb_i$:
\begin{align}\label{eq:classifier}
h(\xb_i) & = \argmax_{k=1,\dots,K} \pb_{k}(\xb_i,\btheta), \\
\pb_k(\xb_i,\btheta) & = \exp(\zb_{k}(\xb_i,\btheta)) / \sum_{k'=1}^K\exp(\zb_{k'}(\xb_i,\btheta)),
\end{align}
where $\zb_{k}(\xb_i,\btheta)$ is the logits output of the network with respect to class $k$, and $\pb_{k}(\xb_i,\btheta)$ is the probability (softmax on logits) of $\xb_i$ belonging to class $k$. The model parameters $\btheta$ are updated using back-propagation to minimize the classification loss such as the commonly used cross entropy loss $\ell(h, \xb)=\frac{1}{N}\sum_i^N - y_i \log \pb_{y_i}(\xb_i,\btheta)$.

\subsection{Adversarial Attacks.}
Given a pretrained DNN model $h$ and a normal sample $\xb$ with class label $y$, an attacking method is to maximize the classification error of the DNN model, whilst keeping $\xa$ within a small $\epsilon$-ball centered at the original sample $\xb$ ($\|\xa - \xb\|_p \leq \epsilon$), where $\|\cdot\|_p$ is the $L_p$-norm, with $L_{\infty}$ being the most commonly used norm due to its consistency with respect to human perception \cite{madry2018towards}. 
Adversarial attacks can be either targeted or untargeted. A targeted attack is to find an adversarial example $\xa$ that can be predicted by the DNN to a target class ($h(\xa) = y_{target}$) which is different from the true class ($y_{target} \neq y$), while an untargeted attack is to find an adversarial example $\xa$ that can be misclassified to an arbitrary class ($h(\xa) \neq y$). Adversarial attacks can be generated either in a white-box setting using adversarial gradients extracted directly from the target model, or a black-box setting by attacking a surrogate model or estimation of the adversarial gradients \cite{jiang2019black,Wu2020Skip}. In this paper, we focus on untargeted attacks in the white-box setting under the $L_{\infty}$ perturbation constraint.

For white-box untargeted attacks, adversarial examples can be generated by solving the following constrained optimization problem:
\begin{equation}\label{eq:prob}
   \xa = \argmax_{\|\xb' - \xb\|_\infty \leq \epsilon} \ell(h(\xb'), y),
\end{equation}
where $\ell(\cdot)$ is the classification loss, and $y$ is the ground truth class. A wide range of attacking methods have been proposed for the crafting of adversarial examples. Here, we introduce a selection of the most representative and state-of-the-art attacks.

\textbf{Fast Gradient Sign Method (FGSM).} FGSM perturbs normal examples $\xb$ for one step by the amount of $\epsilon$ along the input gradient direction \cite{goodfellow2014explaining}:
\begin{equation}
    \xa = \xb + \epsilon \cdot \text{sign}(\nabla_{\xb} \ell(h(\xb), y)).
\end{equation}

\textbf{Basic Iterative Method (BIM).} BIM  \cite{kurakin2017adversarial} is an iterative version of FGSM. Different to FGSM, BIM iteratively perturbs the input with smaller step size,
\begin{equation}
\xb^{t} = \big( \xb^{t-1} + \alpha \cdot \text{sign}(\nabla_{\xb} \ell(h(\xb^{t-1}), y)),
\end{equation}
where $\alpha$ is the step size, and $\xb^{t}$ is the adversarial example at the $t$-th step ($\xb^{0} = \xb$). 
The step size is usually set to $\epsilon/T \leq \alpha  < \epsilon$ for overall $T$ steps of perturbation.

\textbf{Projected Gradient Descent (PGD).} PGD \cite{madry2018towards} perturbs a normal example $\xb$ for a number of $T$ steps with smaller step size. After each step of perturbation, PGD projects the adversarial example back onto the $\epsilon$-ball of $\xb$, if it goes beyond:
\begin{equation}
\xb^{t} = \Pi_{\epsilon} \big( \xb^{t-1} + \alpha \cdot \text{sign}(\nabla_{\xb} \ell(h(\xb^{t-1}), y)) \big),
\end{equation}
where $\alpha$ is the step size, $\Pi(\cdot)$ is the projection function, and $\xb^{t}$ is the adversarial example at the $t$-th step ($\xb^{0} = \xb$). Different from BIM, PGD uses random start for $\xb^{0}= \xb + \cU^{d}(-\epsilon, \epsilon)$, where $\cU^{d}(-\epsilon, \epsilon)$ is the uniform distribution between $-\epsilon$ and $\epsilon$, and of the same $d$ dimensions as $\xb$. PGD is normally regarded as the strongest first-order attack.

\textbf{Carlini and Wagner (CW) Attack}. 
The CW attack is a state-of-the-art optimization-based attack \cite{carlini2017towards}. There are two versions of the CW attack: $L_2$ and $L_{\infty}$, here we focus on the $L_{\infty}$ version. According to \cite{madry2018towards}, the $L_{\infty}$ version of targeted CW attack can be solved by the PGD algorithm iteratively as following
\begin{align}
    \xb^{t} &= \Pi_{\epsilon} \big( \xb^{t-1} - \alpha \cdot \text{sign}(\nabla_{\xb} \hat{f}(\xb^{t-1})) \big)\\
\hat{f}(\xb^{t-1}) &= \max \big(\zb_{y}(\xb^{t-1}, \btheta) - \zb_{y_{max} \neq y}(\xb^{t-1}, \btheta), - \kappa \big),
\end{align}
where $\hat{f}(\cdot)$ is the surrogate loss for the constrained optimization problem defined in Eqn. \eqref{eq:prob}, $\zb_y$ is the logits with respect to class $y$, $\zb_{y_{max} \neq y}$ is the maximum logits of other classes, and $\kappa$ is a parameter controls the confidence of the attack.

While there also exists other attacking methods \cite{Wu2020Skip}, in this paper, we focus on the four state-of-the-art attacks mentioned above: FGSM, BIM, PGD and CW.

\subsection{Adversarial Detection}
A number of defense models have been developed,
input denoising \cite{bai2019hilbert}, input gradients regularization \cite{ross2018improving}, and adversarial training \cite{goodfellow2014explaining, madry2018towards}. However, these defenses can generally be evaded by the latest attacks, either wholly or partially \cite{athalye2018obfuscated}.

Given the inherent challenges for adversarial defense, recent works have instead focused on detecting adversarial examples. These works attempt to discriminate adversarial examples (positive class) from normal clean examples (negative class), based on features extracted from different layers of a DNN. 
In machine learning, the subspace distance of the high dimension features has long been analysed \cite{Zhou_PR20}.
Specifically, for the adversarial examples detection,  detection subnetworks based on activations \cite{metzen2017detecting}, 
a logistic regression detector based on KD and Bayesian Uncertainty (BU) features \cite{feinman2017detecting} and the Local Intrinsic Dimensionality (LID) of adversarial subspaces \cite{ma2018characterizing} are a few such works. 

\textbf{Kernel Density (KD):} KD assumes that normal samples from the same class lie densely on the data manifold while adversarial samples lie in more sparse regions off the data submanifold. Given a point $\xb$ of class $k$, and a set of training samples from the same class $X_k$, the Gaussian Kernel Density of $\xb$ can be estimated by:
\begin{equation}
\label{eq:kd}
  \text{KD}(\xb) = \frac{1}{|X_k|} \sum_{\xb' \in X_k} \text{exp}\big(\frac{\rVert\zb(\xb, \btheta)- \zb(\xb', \btheta)\rVert_2^2}{\sigma^2}\big),
\end{equation}
where $\sigma$ is the bandwidth parameter controlling the smoothness of the Gaussian estimation, $\zb$ is the logits of input $\xb$, and $|X_k|$ is the number of samples in $X_k$.

\textbf{Local Intrinsic Dimensionality (LID):} LID is a measurement to characterize the dimensional characteristics of adversarial subspaces in the vicinity of adversarial examples. Given an input sample $\xb$, the MLE estimator of LID makes use of its distances to the first $n$ nearest neighbors:
\begin{equation} \label{eq:estimator}
\widehat{\textup{LID}}(\xb) = - \Bigg( \frac{1}{n}\sum_{i=1}^{n}\log \frac{r_i(\xb)}{r_n(\xb)}\Bigg)^{-1},
\end{equation}
where $r_i(\xb)$ is the Euclidean distance between $\xb$ and its $i$-th nearest neighbor, i.e, $r_1(\xb)$ is the minimum distance while $r_n(\xb)$ is the maximum distance. LID is computed on each layer of the network producing a vector of LID scores for each sample.

\subsection{Classification Tasks, Datasets and DNN Models}\label{sec:experiments}
Here, we consider three highly successful applications of DNNs for medical image classification: 1) classifying diabetic retinopathy (a type of eye disease) from retinal fundoscopy \cite{gulshan2016development}; 2) classifying thorax diseases from Chest X-rays \cite{wang2017chestxray}; and 3)
classifying melanoma (a type of skin cancer) from dermoscopic photographs \cite{esteva2017dermatologist}. 
Here, we briefly introduce some general experimental settings with respect to the datasets and network architectures.

\textbf{Datasets.}
We use publicly available benchmark datasets for all three classification tasks. For our model training and attacking experiments, we need two subsets of data for each dataset: 1) subset \emph{Train} for pre-training the DNN model, and 2) subset \emph{Test} for evaluating the DNN models and crafting adversarial attacks. In the detection experiments, we further split the \emph{Test} data into two parts: 1) \emph{AdvTrain} for training adversarial detectors, and 2) \emph{AdvTest} for evaluating the adversarial detectors. The number of classes and images we retrieved from the public datasets can be found in Table~\ref{table:dataset}.

\begin{table}[!htb]
\caption{Number of classes and images in each subset of the five datasets.}
\label{table:dataset}
\centering
\small{
    \begin{tabular}{c|c|c|cc}
    \toprule
    \multirow{2}[2]{*}{Dataset} &
    \multirow{2}[2]{*}{Classes} & \multirow{2}[2]{*}{\emph{Train}} & \multicolumn{2}{c}{\emph{Test}} \\
    & &  & \emph{AdvTrain} & \emph{AdvTest}\\
    \midrule
    Fundoscopy & 2 & 75,397 & 8,515 & 2,129 \\
    Chest X-Ray & 2 & 53,219 & 6,706 & 1,677 \\
    Dermoscopy & 2 & 18,438 & 426 & 107 \\
    \midrule
    Chest X-Ray-3 & 3 & 54769 & \multicolumn{2}{c}{9980} \\
    Chest X-Ray-4 & 4 & 57059 & \multicolumn{2}{c}{10396}\\
    \bottomrule
    \end{tabular}}
\end{table}

We follow the data collection process described in \cite{finlayson2019adversarial}. For the diabetic retinopathy (DR) classification task, we use the Kaggle dataset Fundoscopy \cite{kaggle2015diabetic}, which consists of over 80,000 high-resolution retina images taken under a variety of imaging conditions where each image was labeled to five scales from `No DR' to `mid/moderate/severe/proliferative DR'. In accordance with \cite{gulshan2016development,finlayson2019adversarial}, we aim to detect the \textit{referable} (grade moderate or worse) diabetic retinopathy from the rest (two classes in total).

For the thorax disease classification task, we use a Chest X-Ray database~\cite{wang2017chestxray}, which comprises 112,120 frontal-view X-ray images of 14 common disease labels. Each image in this dataset can have multiple labels, so we randomly sample images from those labeled only with `no finding' or `pneumothorax' to obtain our 2-class dataset.
We also sample two multi-class datasets from Chest X-Ray: 1) a 3-class dataset (eg. Chest X-Ray-3 in Table \ref{table:dataset}) including image labeled only with `no finding', `pneumothorax' or `mass'; 2) a 4-class dataset (eg. Chest X-Ray-4 in Table \ref{table:dataset}) including `no finding', `pneumothorax', `mass' and `nodule'. 

\begin{figure*}[!ht]
    \centering
    \includegraphics[width=0.8\textwidth]{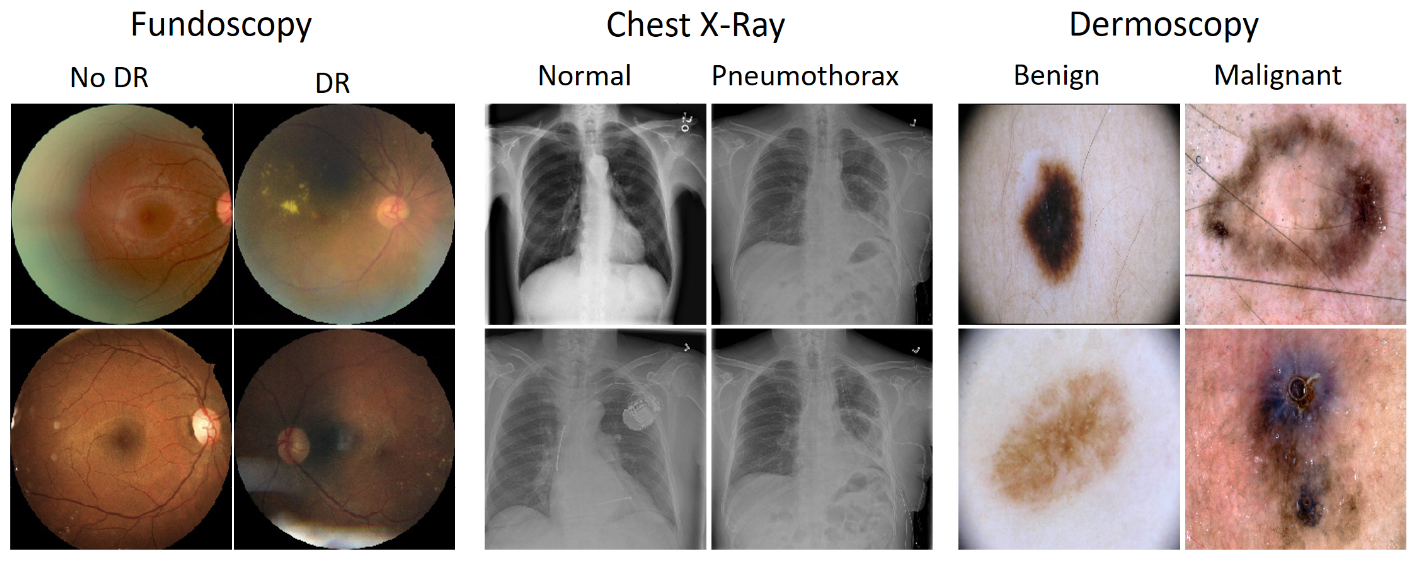}
    \caption{Example images from each class of the three 2-class datasets.}
    \label{fig:datasets}
\end{figure*}

For the melanoma classification task, we retrieve melanoma related images of class `benign' and class `malignant' (two classes in total) from the International Skin Imaging Collaboration database \cite{isic_archive}. Figure \ref{fig:datasets} shows two examples for each class of our three 2-class datasets.

\textbf{DNN Models.}
For all the five datasets, we use the ImageNet pretrained ResNet-50 \cite{he2016deep} as the base network whose top layer is replaced by a new dense layer of 128 neurons, followed by a dropout layer of rate 0.2, and a $K$ neuron dense layer for classification. The networks are trained for 300 epochs using a stochastic gradient descent (SGD) optimizer with initial learning rate $10^{-4}$, momentum 0.9. All images are center-cropped to the size $224\times224\times3$ and normalized to the range of $[-1,1]$. Simple data augmentations including random rotations, width/height shift and horizontal flip are used. When the training is completed, the networks are fixed in subsequent adversarial experiments.

\section{Understanding Adversarial Attacks on Medical Image DNNs}\label{sec:attack}
In this section, we investigate 4 different attacks against DNNs trained on five medical image datasets. We first describe the attack settings, then present the attack results with accompanying discussions and analyses.

\begin{figure}[!t]
    \centering
    \includegraphics[width=\columnwidth]{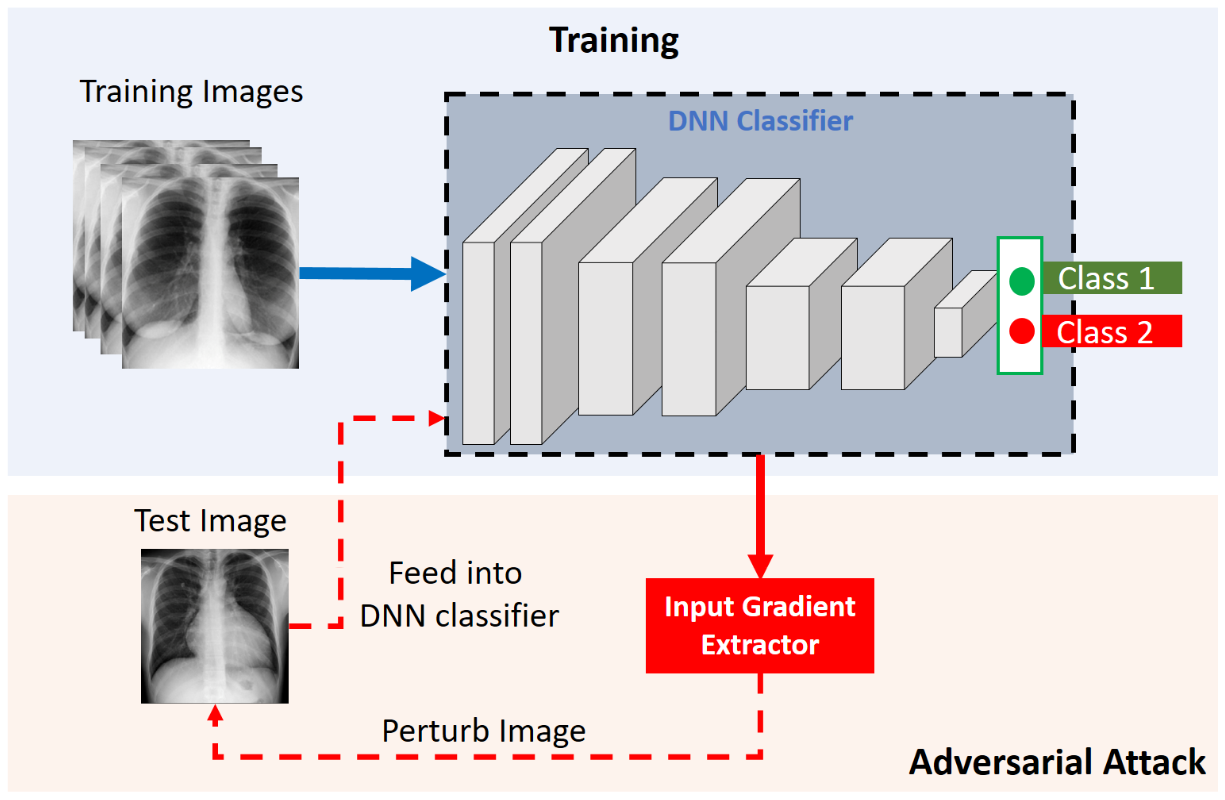}
    \caption{The pipeline of training DNNs (top) and generating adversarial attacks (bottom).}
    \label{fig:attack_pipeline}
\end{figure}

\subsection{Attack Settings}
The attacks we consider are: 1) the single step attack FGSM, 2) the iterative attack BIM, 3) the strongest first-order attack PGD, and 4) the strongest optimization-based attack CW ($L_{\infty}$ version). Note that all these attacks are bounded attacks according to a pre-defined maximum perturbation $\epsilon$ with respect to the $L_{\infty}$ norm, \textit{i.e.}, the maximum perturbation on each input pixel is no greater than $\epsilon$. All 4 types of attacks are applied on both the \emph{AdvTrain} and \emph{AdvTest} subsets of images, following the pipeline in Figure \ref{fig:attack_pipeline}. Given an image, the input gradient extractor feeds the image into the pre-trained DNN classifier to obtain the input gradients, based upon which the image is perturbed to maximize the network's loss to the correct class.  The perturbation steps for BIM, PGD and CW are set to 40, 20 and 20 respectively, while the step size are set to $\epsilon/40$, $\epsilon/10$ and $\epsilon/10$ accordingly.
We focus on untargeted attacks in a white-box setting.

\begin{figure*}[!ht]
\centering
\begin{subfigure}{.22\linewidth}
  \centering
  \includegraphics[width=\textwidth]{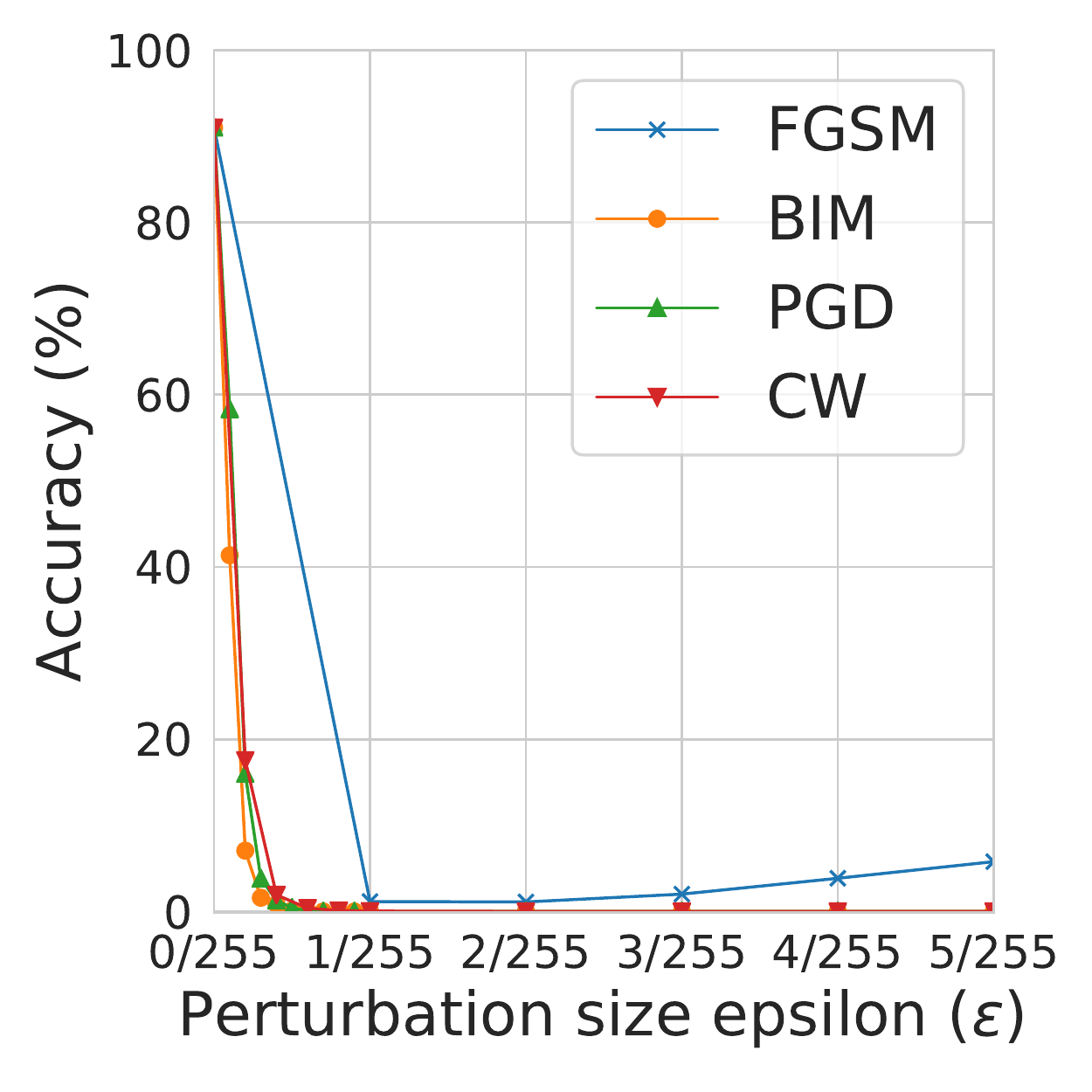}
  \caption{Fundoscopy}
  \label{fig:advs_fundo}
\end{subfigure}
\begin{subfigure}{.22\linewidth}
  \centering
  \includegraphics[width=\textwidth]{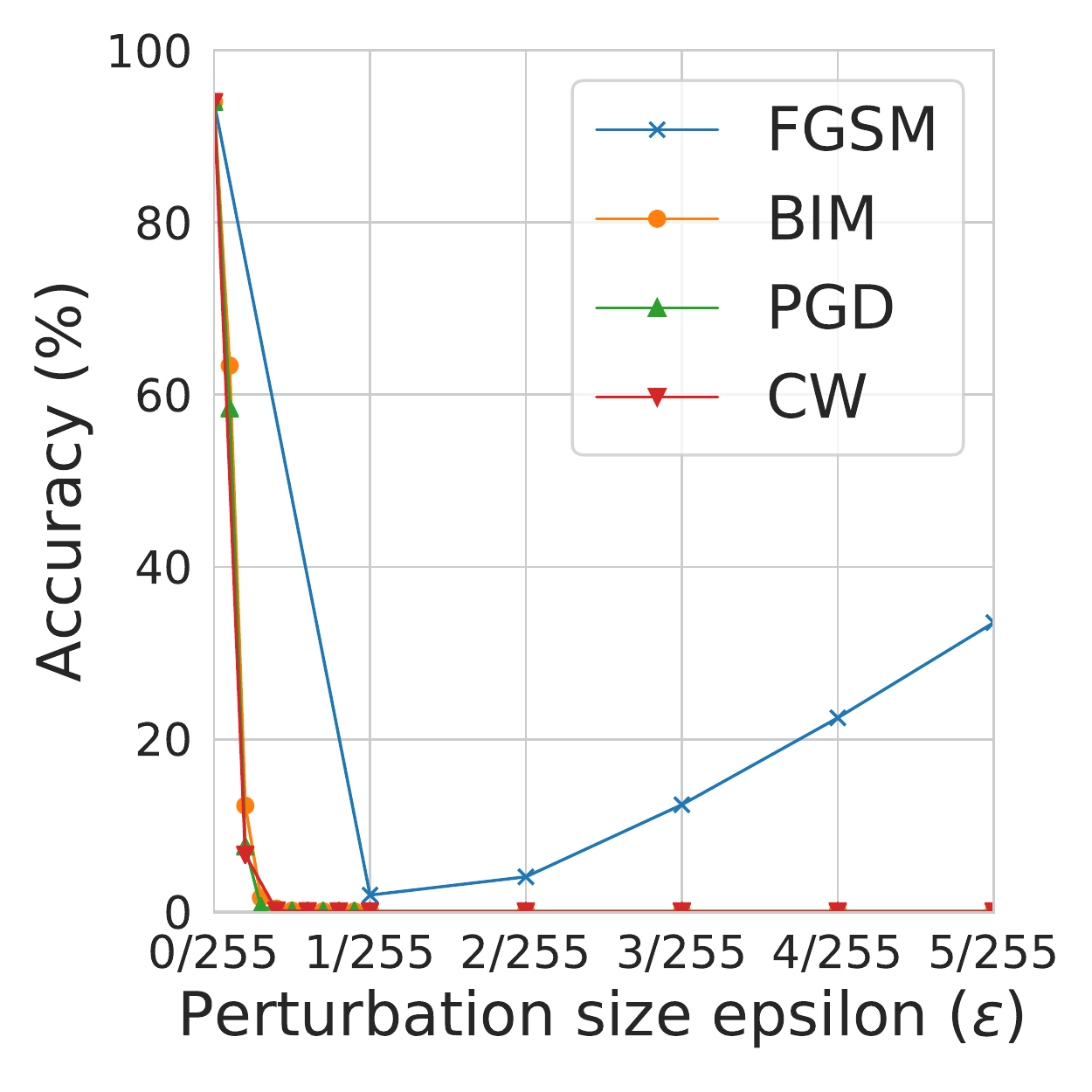}
  \caption{Chest X-Ray}
  \label{fig:advs_chest}
\end{subfigure}
\begin{subfigure}{.22\linewidth}
  \includegraphics[width=\textwidth]{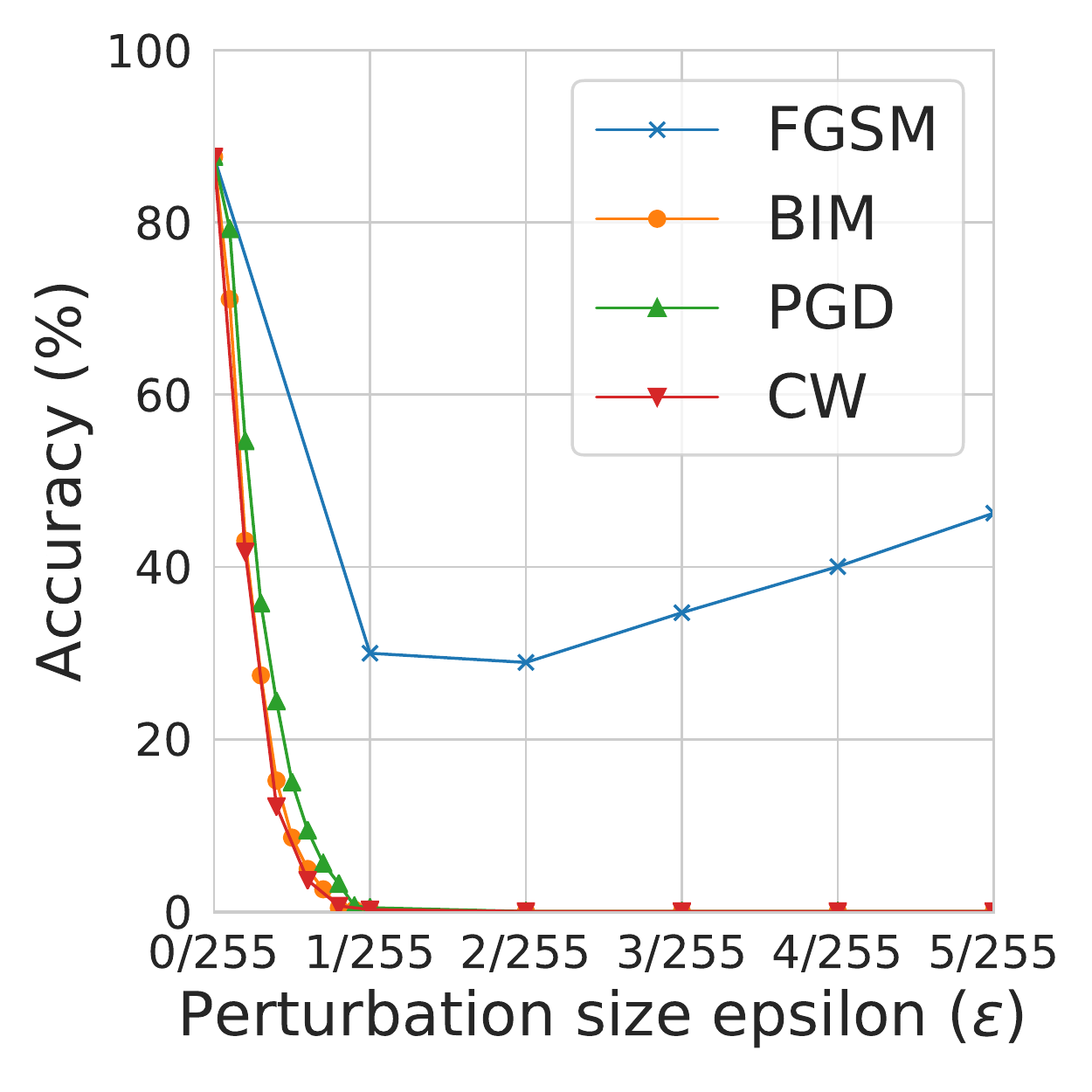}
  \caption{Dermoscopy}
  \label{fig:advs_derm}
\end{subfigure}
\caption{The classification accuracy of the three 2-class DNN classifiers on adversarial examples crafted by FGSM, BIM, PGD and CW with increasing perturbation size $\epsilon$. Strong attacks including BIM, PGD and CW can succeed most of the time (model accuracy below $1\%$) with very small perturbation $< 1.0/255$. All attacks were generated in a white-box setting.}
\label{fig:adv_attack}
\end{figure*}

\begin{table*}[!hbt]
    \centering
    \begin{tabular}{c|cc|cc|cc}
    \toprule
    \multirow{2}[2]{*}{\hspace{2em}Attack\hspace{2em}} & \multicolumn{2}{c|}{\hspace{2em}Fundoscopy\hspace{2em}} & \multicolumn{2}{c|}{\hspace{2em}Chest X-Ray\hspace{2em}} & \multicolumn{2}{c}{\hspace{2em}Dermoscopy\hspace{2em}} \\
          & Accuracy & AUC   & Accuracy & AUC   & Accuracy & AUC \\
    \midrule
    No attack & 91.03  & 81.91  & 93.99  & 61.25  & 87.62  & 78.74  \\
    FGSM  & 1.15  & 3.71  & 1.90  & 0.96  & 29.98  & 20.58  \\
    BIM   & 0.00  & 0.00  & 0.00  & 0.00  & 0.21  & 0.13  \\
    PGD   & 0.00  & 0.00  & 0.00  & 0.00  & 0.43  & 0.74  \\
    CW    & 0.04  & 0.09  & 0.00  & 0.00  & 0.21  & 0.13  \\
    \bottomrule
    \end{tabular}%
	\caption{The classification accuracies (\%) and AUCs (\%) of the three 2-class DNN classifiers on clean test images (denoted as ``No attack") and the 4 types of adversarial examples under $L_{\infty}$ maximum perturbation $1.0/255$.}
	\label{tab:attack_white}
\end{table*}

\subsection{Attack Results}
\begin{figure*}[!htb]
\centering
\begin{subfigure}{.22\linewidth}
  \centering
  \includegraphics[width=\textwidth]{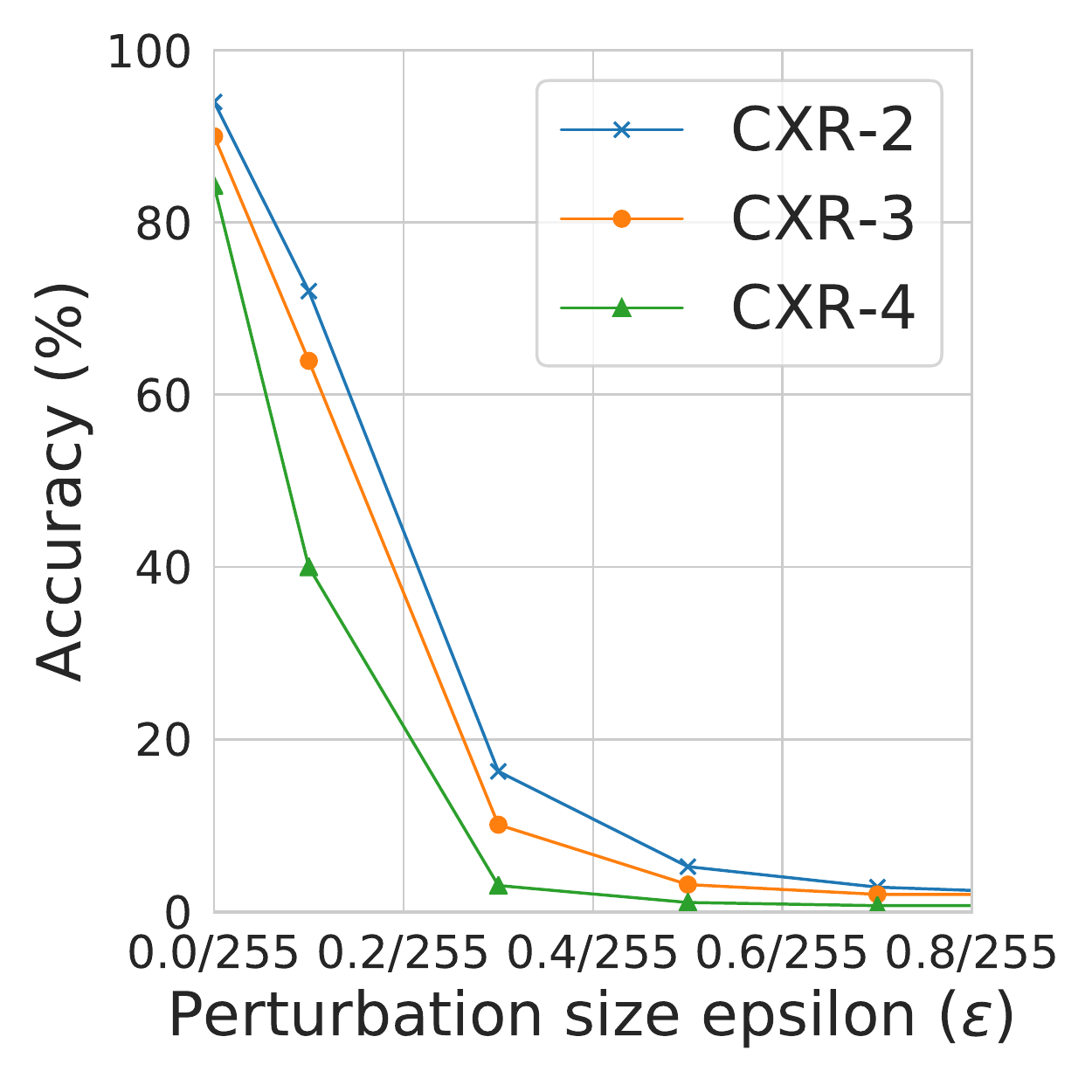}
  \caption{FGSM}
  \label{fig:advs_fgsm}
\end{subfigure}
\begin{subfigure}{.22\linewidth}
  \centering
  \includegraphics[width=\textwidth]{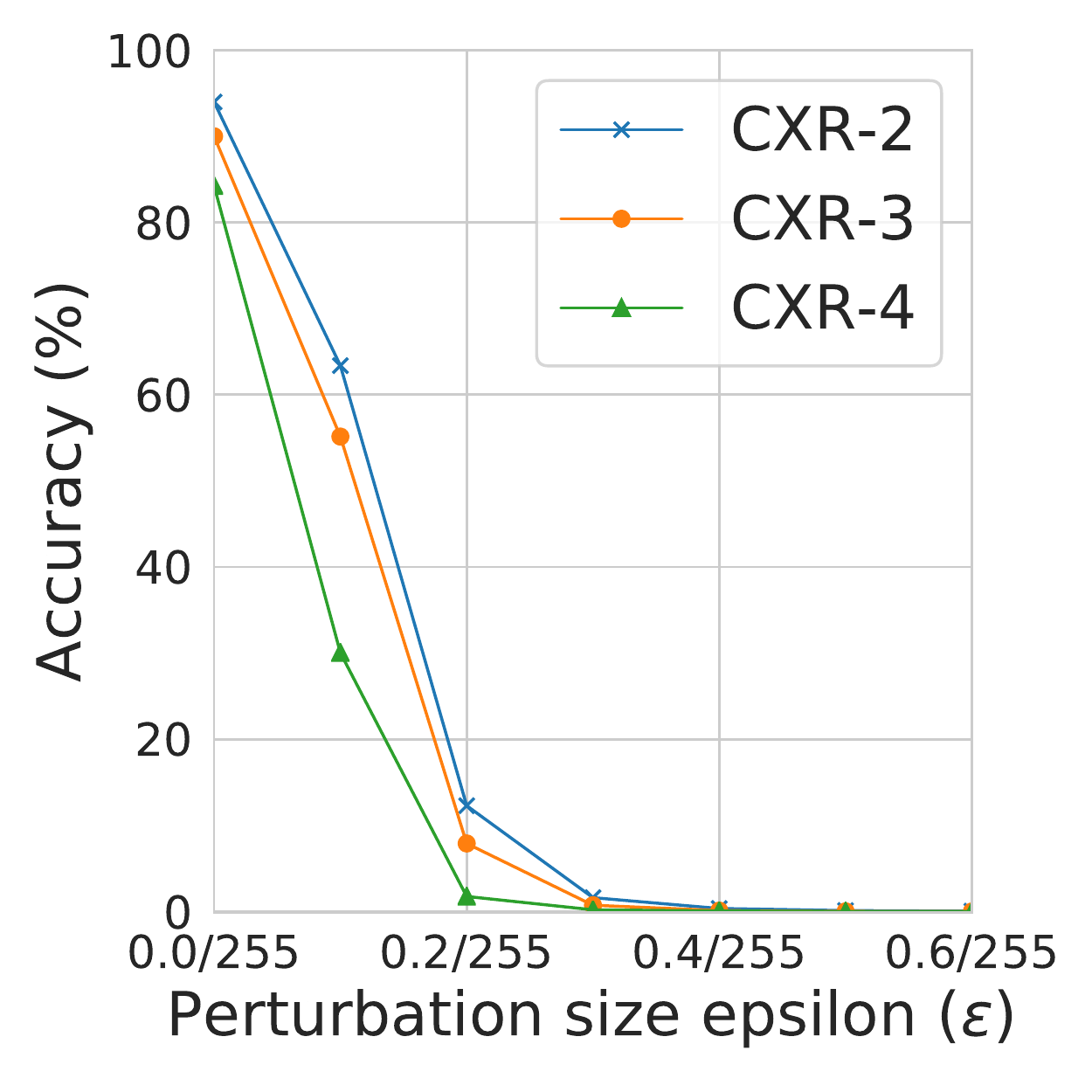}
  \caption{BIM}
  \label{fig:advs_bim}
\end{subfigure}
\begin{subfigure}{.22\linewidth}
  \includegraphics[width=\textwidth]{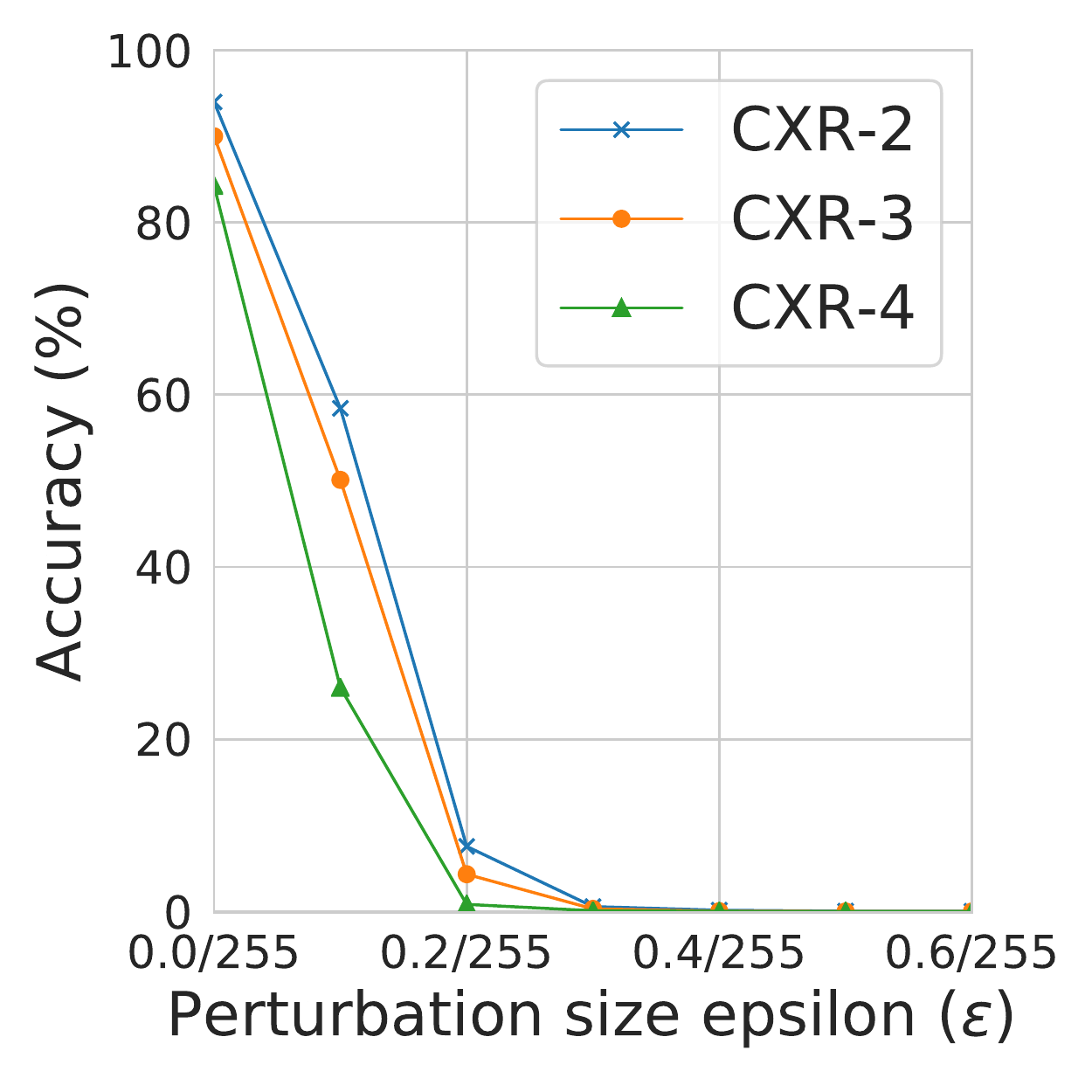}
  \caption{PGD}
  \label{fig:advs_pgd}
\end{subfigure}
\begin{subfigure}{.22\linewidth}
  \includegraphics[width=\textwidth]{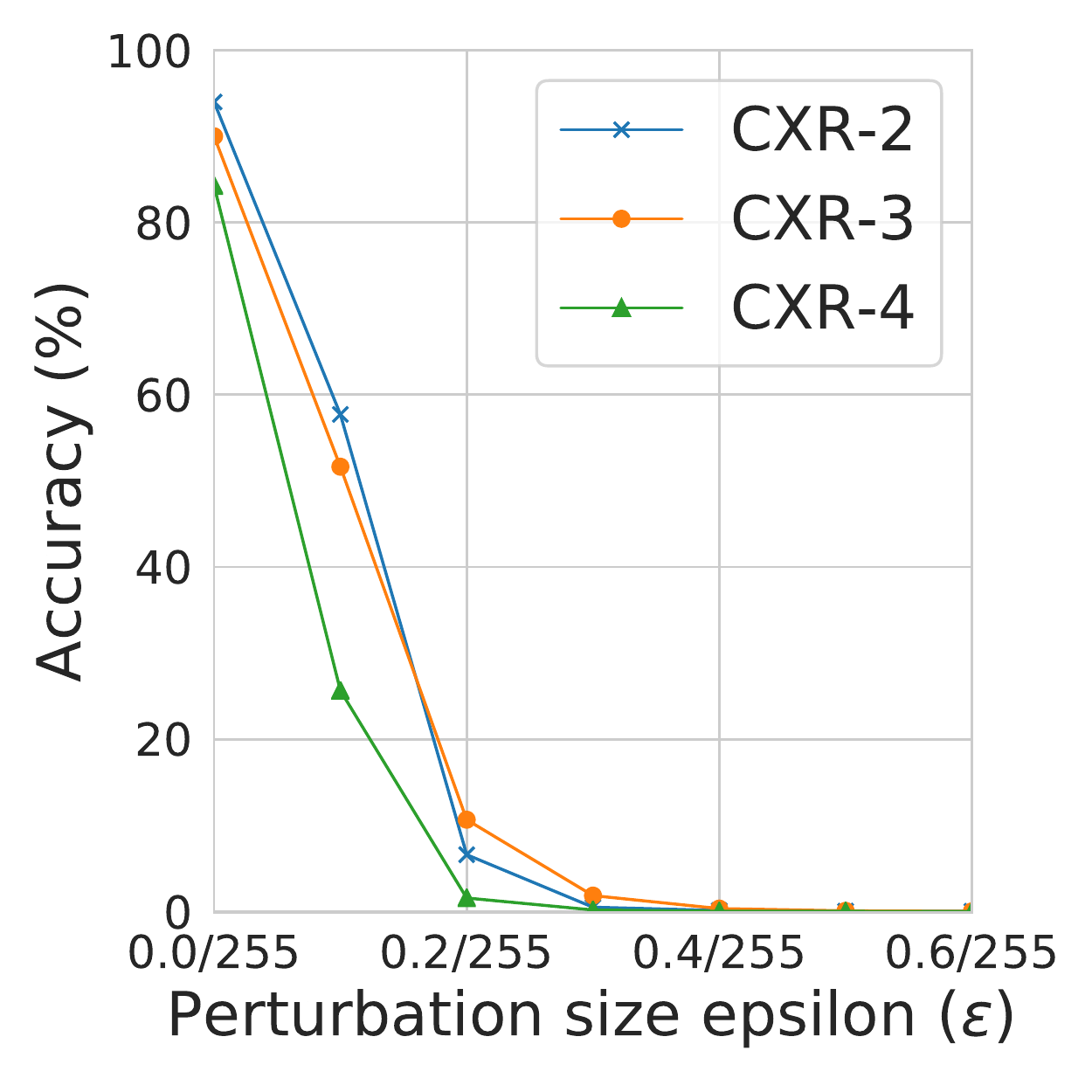}
  \caption{CW}
  \label{fig:advs_cw}
\end{subfigure}
\caption{Comparison of the attacks FGSM, BIM, PGD and CW on datasets Chest X-Ray (CXR-2), Chest X-Ray-3 (CXR-3) and Chest X-Ray-4 (CXR-4). For each attack, the classification accuracy after the attack (in a white-box setting) under different perturbation sizes $\epsilon$ is reported.}
\label{fig:adv_attack2}
\end{figure*}

We focus on the difficulty of adversarial attack on medical images compared to that on natural images in ImageNet. The attack difficulty is measured by the least maximum perturbation required for most (\textit{e.g.} $>99\%$) attacks to succeed. Specifically, we vary the maximum perturbation size $\epsilon$ from 0.2/255 to 5/255, and visualize the drop in model accuracy on the adversarial examples in Figure~\ref{fig:adv_attack} and Figure~\ref{fig:adv_attack2} for our 2-class and multi-class datasets respectively, and the numeric results with respect to maximum perturbation $\epsilon=1.0/255$ can be found in Table \ref{tab:attack_white} and Table \ref{tab:attack_white2} separately.

\begin{table*}[!hbt]
    \centering
    \begin{tabular}{c|ccc|ccc}
    \toprule
    \multirow{2}[2]{*}{Attack} & \multicolumn{3}{c|}{Accuracy when $\epsilon=0.3/255$} & \multicolumn{3}{c}{Accuracy when $\epsilon=1.0/255$} \\
          & CXR-2 & CXR-3 & CXR-4 & CXR-2 & CXR-3 & CXR-4 \\
    \midrule
    No attack & 93.99  & 90.01  & 84.26  & 93.99  & 90.01  & 84.26  \\
    FGSM  & 16.26  & 10.07  & 3.01  & 1.90  & 2.14  & 0.74  \\
    BIM   & 1.60  & 0.72  & 0.19  & 0.00     & 0.00     & 0.00 \\
    PGD   & 0.56  & 0.30  & 0.08  & 0.00     & 0.00     & 0.00 \\
    CW    & 0.49  & 1.84  & 0.17  & 0.00     & 0.00     & 0.00 \\
    \bottomrule
    \end{tabular}%
	\caption{White-box attacks on 2-class versus multi-class models on datasets Chest X-Ray (CXR-2), Chest X-Ray-3 (CXR-3) and Chest X-Ray-4 (CXR-4): the classification accuracies (\%) of the three DNN classifiers on clean test images (denoted as ``No attack") and the 4 types of adversarial examples under $L_{\infty}$ maximum perturbation $\epsilon=0.3/255$ and $\epsilon=1.0/255$.}
	\label{tab:attack_white2}
\end{table*}

\noindent\textbf{Results on 2-class datasets.}
As expected, model accuracy drops drastically when adversarial perturbation increases, similar to that on natural images \cite{goodfellow2014explaining,carlini2017towards}.
Strong attacks including BIM, PGD and CW, only require a small maximum perturbation $\epsilon < 1.0/255$ to generally succeed. This means attacking medical images is much easier than attacking natural images like those from CIFAR-10 and ImageNet, which often require a maximum perturbation of $> 8.0/255$ for targeted attacks to generally succeed (see Figure 2 in  \cite{kurakin2017adversarial}).

\noindent\textbf{Results on multi-class datasets.}
Here, we further investigate the attack difficulty on 2-class datasets (eg. Chest X-Ray) versus that on multi-class datasets (eg. Chest X-Ray-3 and Chest X-Ray-4). As the AUC score is defined with respect to only 2 classes, here we only report the model accuracy on clean images (eg. ``No attack") and adversarial images crafted by FGSM, BIM, PGD, and CW. As shown in Table \ref{tab:attack_white2}, when there are more classes, the attacks have greater success rate. For example, under the same perturbation $\epsilon=0.3/255$, model accuracy on crafted adversarial examples decreases as the number of classes increases. This indicates that medical image datasets that have multiple classes are even more vulnerable than those 2-class datasets. Similar to the 2-class results above, the attacks BIM, PGD and CW can succeed more than 99\% of the time with small perturbation $\epsilon=1.0/255$. This is the case even with smaller perturbation $\epsilon=0.3/255$, except for the CW attack on Chest X-Ray-3, which succeeds $>98\%$ of the time. These findings are consistent with those found on natural images, that is, defending adversarial attacks on datasets with more classes (eg. CIFAR-100/ImageNet versus MNIST/CIFAR-10) is generally more difficult \cite{shafahi2019adversarial}.

We next consider further why attacking medical images is much easier than attacking ImageNet images.  At first sight it is surprising, since medical images have the same size as ImageNet images. 

\subsection{Why are Medical Image DNN Models Easy to Attack?}
In this part, we provide explanations to the above phenomenon from the following 2 perspectives: 
1) the characteristics of medical images; and 2) the characteristics of DNN models used for medical imaging.

\begin{figure*}[!hbt]
	\centering
	\includegraphics[width=0.9\textwidth]{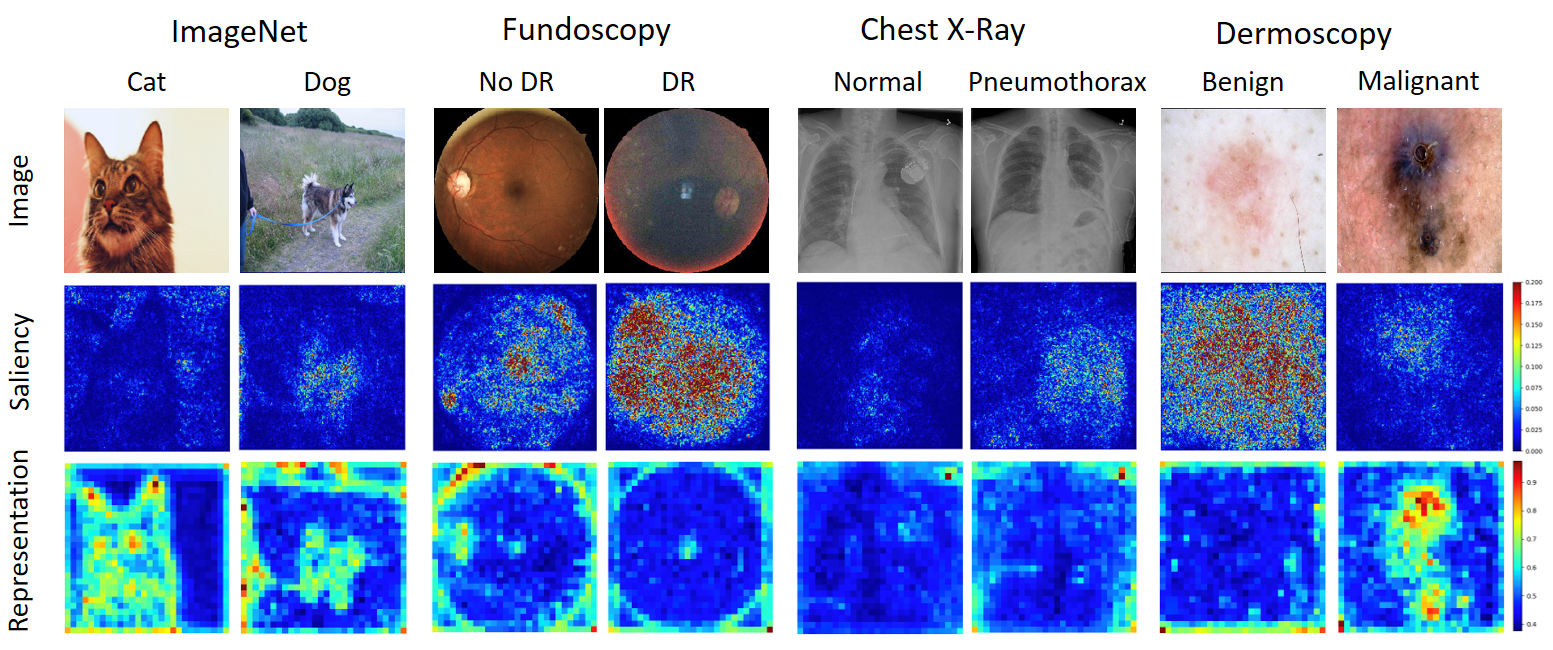}
	\caption{The normal images (top row), the saliency maps of the images (middle row), and their representations (bottom row) learned at the `res3a\_relu' layer (averaged over channels) of the networks. }
	\label{fig:saliency}
\end{figure*}

\textbf{Medical Image Viewpoint.} We show the saliency map for several images from different classes, for both ImageNet and medical images in the middle row of Figure \ref{fig:saliency}. The saliency (or attention) map of an input image highlights the regions that cause the most change in the model output, based on the gradients of the classification loss with respect to the input \cite{simonyan2013deep}. 
We can observe that some medical images have significantly larger high attention regions. This may indicate that the rich biological textures in medical images sometimes distract the DNN model into paying extra attention to areas that are not necessarily related to the diagnosis. Small perturbations in these high attention regions can lead to significant changes in the model output. In other words, this characteristic of medical images increases their vulnerability to adversarial attacks. However, this argument only provides a partial answer to the question, as there is no doubt that some natural images can also have complex textures.

\begin{figure*}[!hbt]
	\centering
	\includegraphics[width=0.9\textwidth]{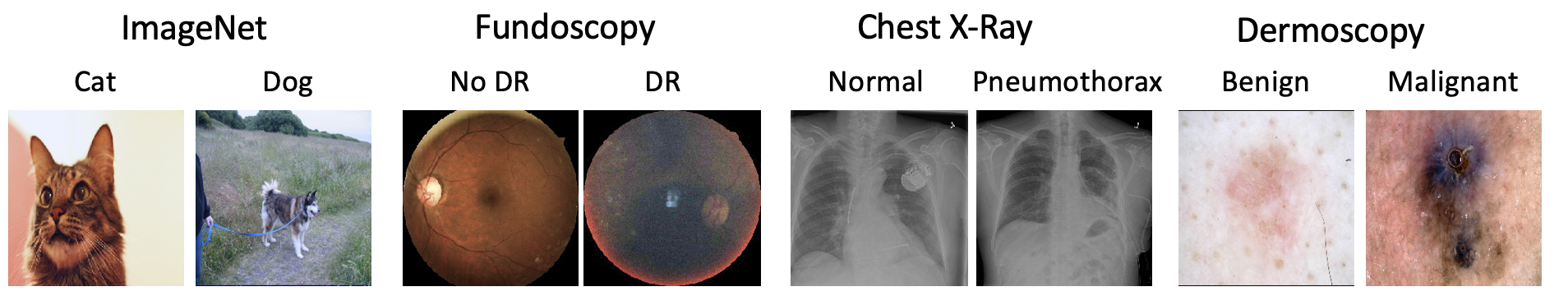}
	\includegraphics[width=0.105\textwidth, trim=90 30 50 50, clip]{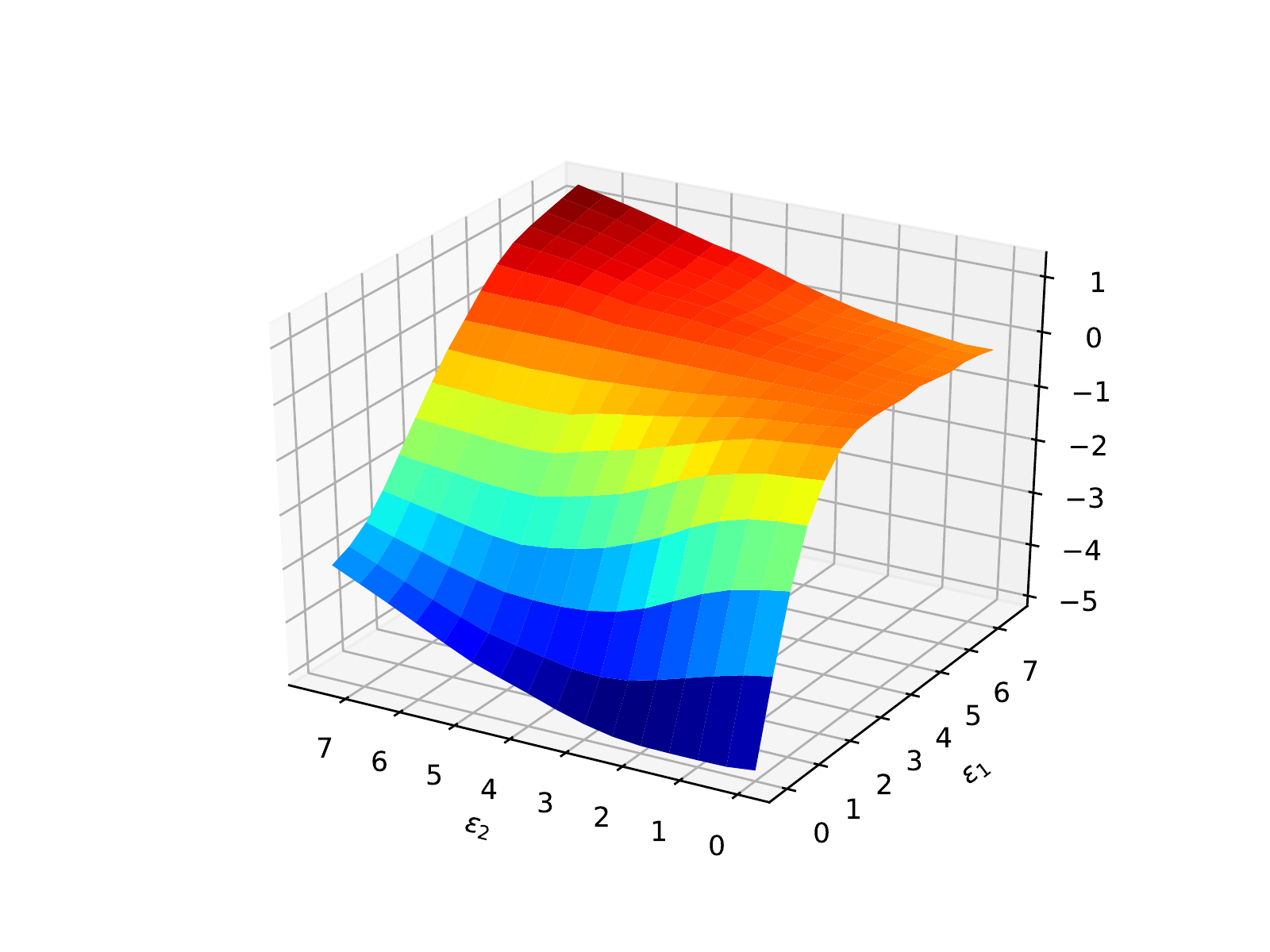} %
	\includegraphics[width=0.105\textwidth, trim=90 30 50 50, clip]{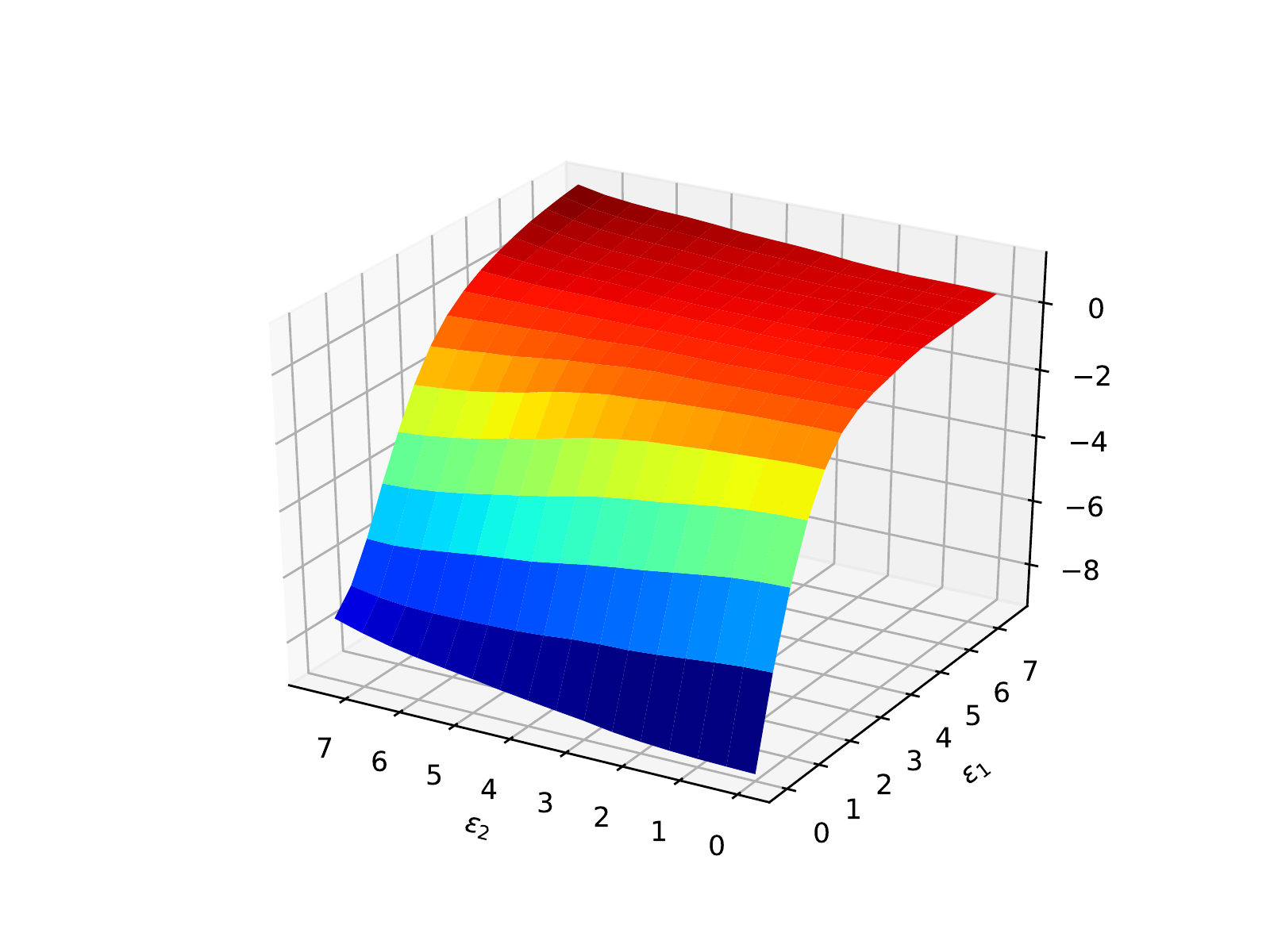} \hspace{0.3mm}
	\includegraphics[width=0.105\textwidth, trim=90 30 50 50, clip]{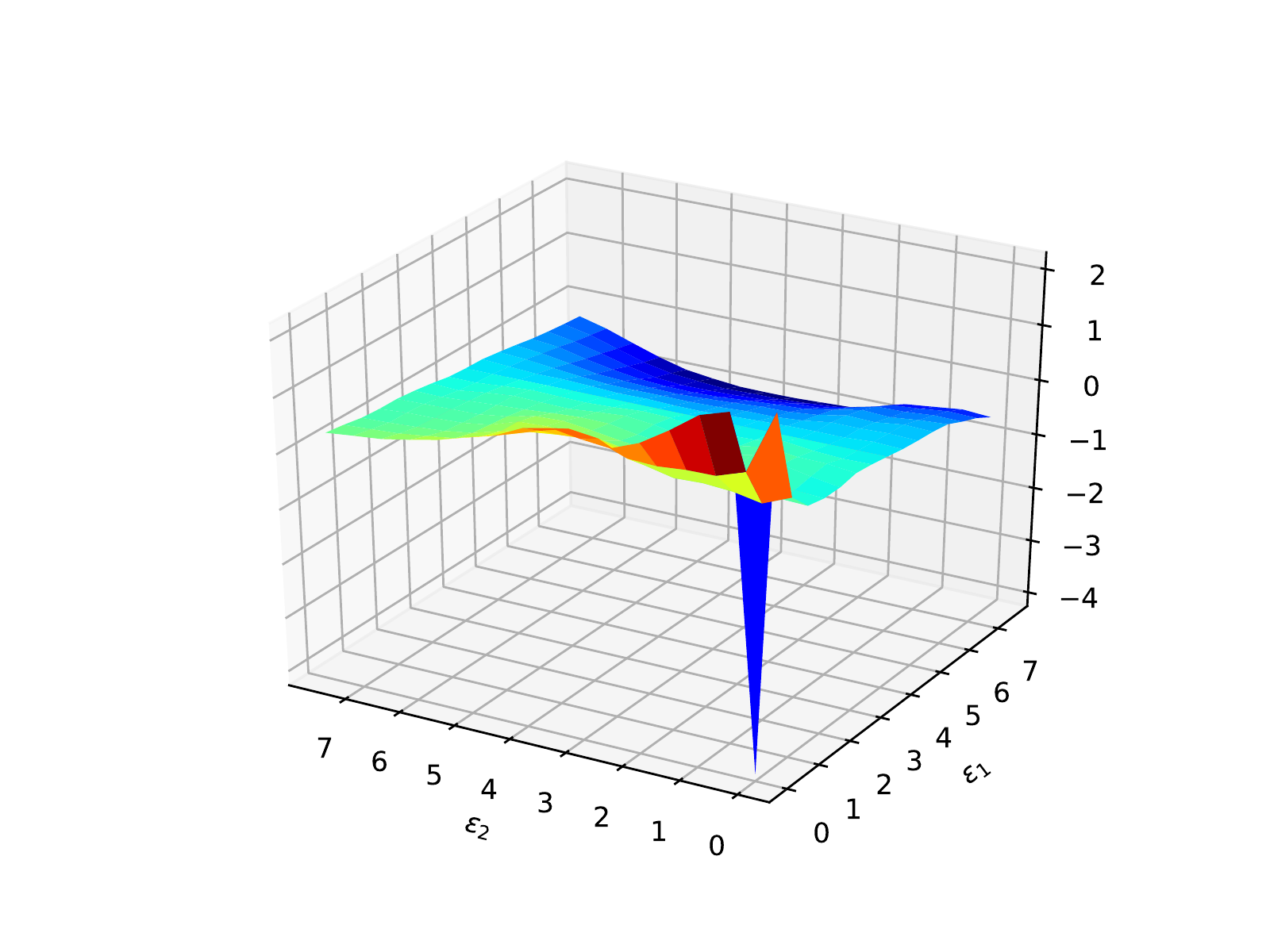} %
	\includegraphics[width=0.105\textwidth, trim=90 30 50 50, clip]{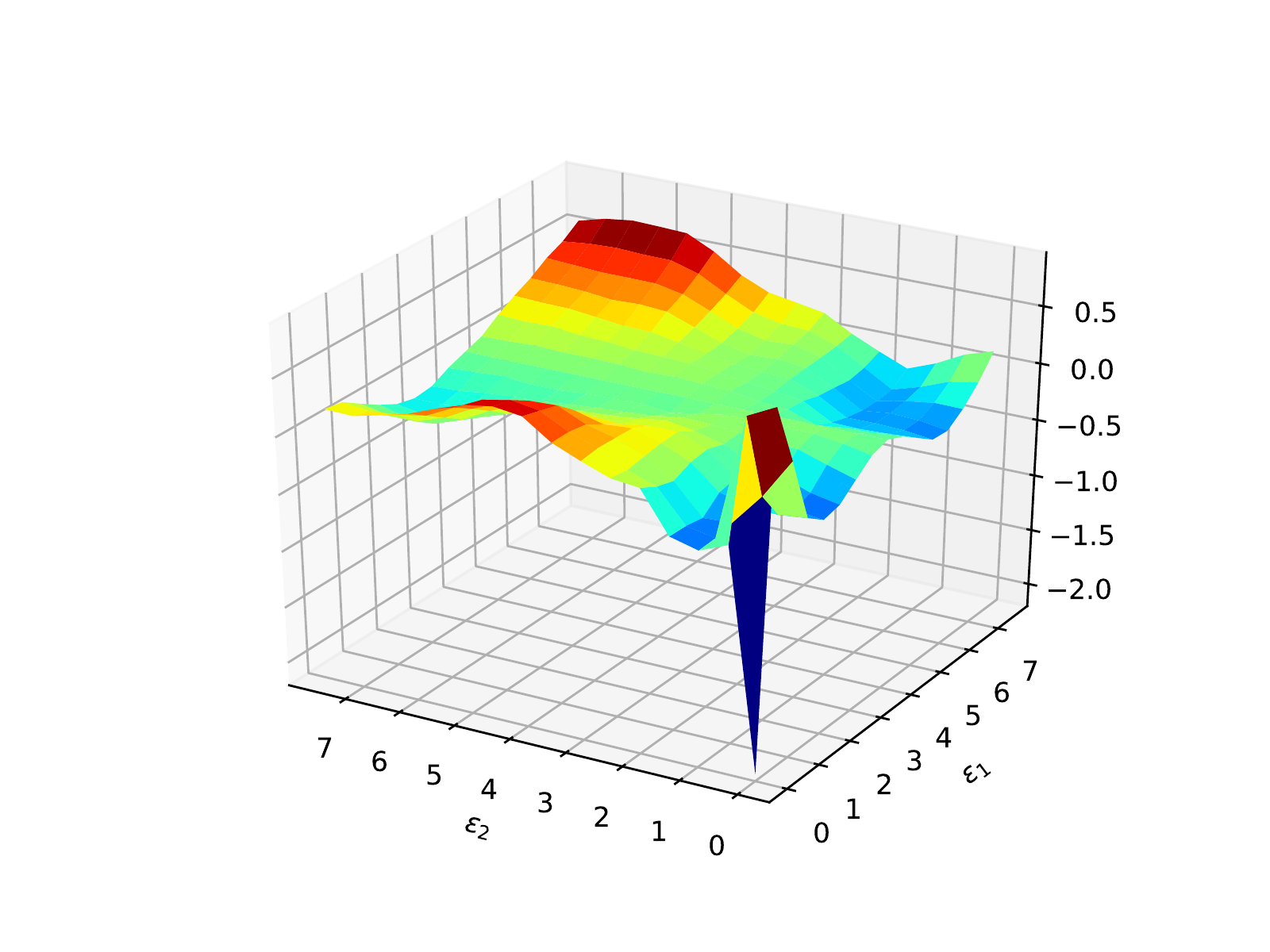} \hspace{0.3mm}
	\includegraphics[width=0.105\textwidth, trim=90 30 50 50, clip]{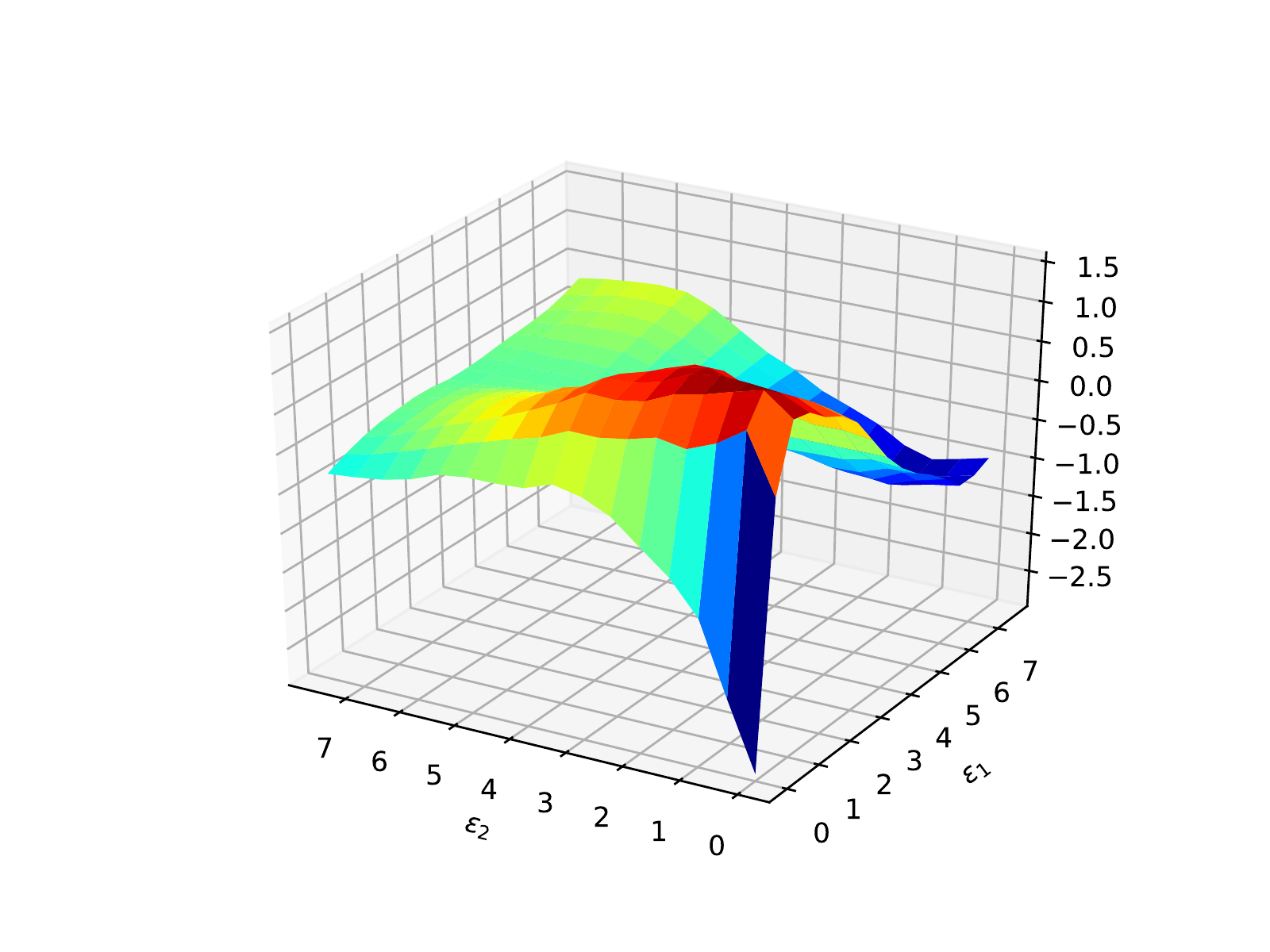} %
	\includegraphics[width=0.105\textwidth, trim=90 30 50 50, clip]{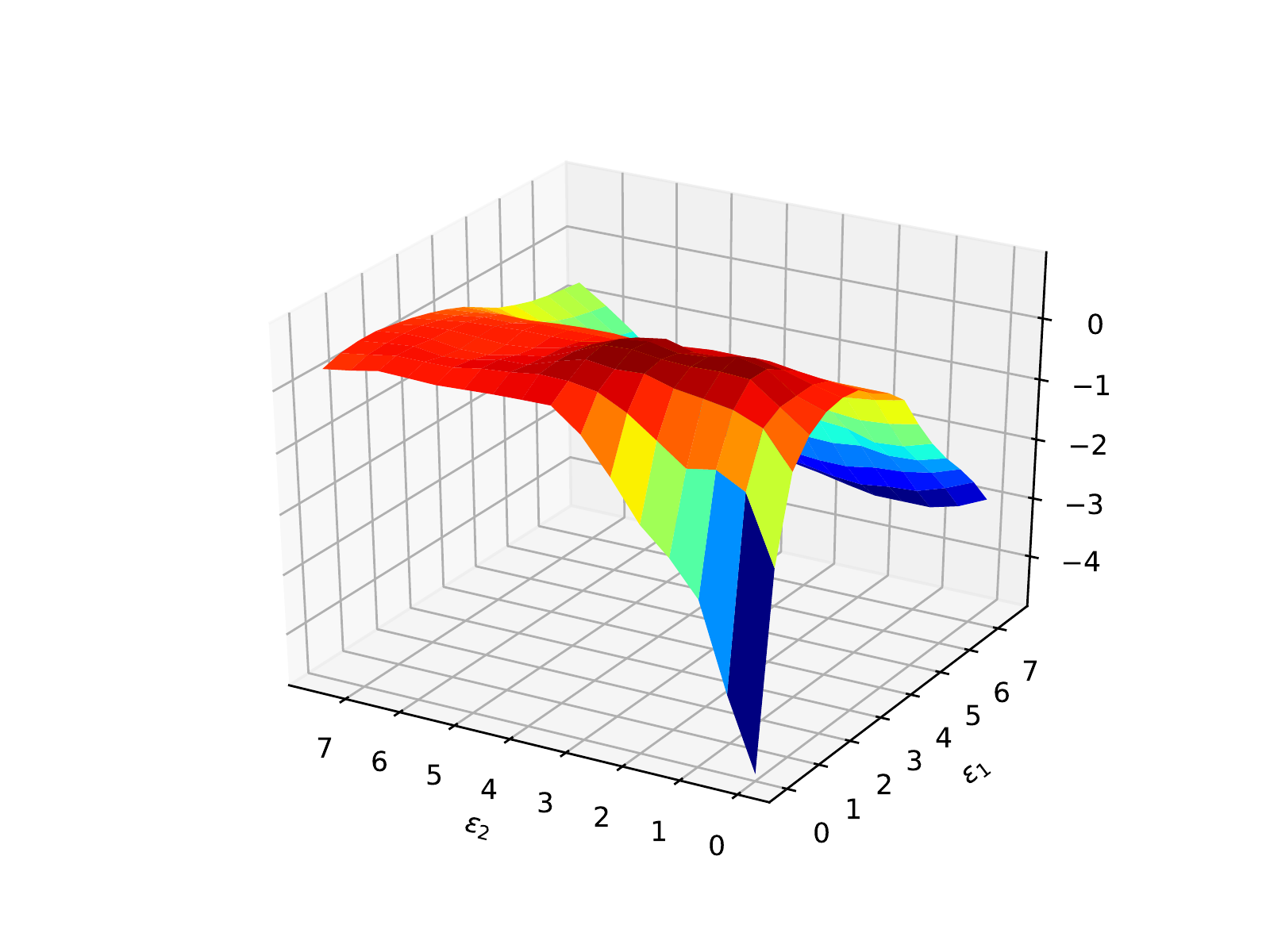} \hspace{0.3mm}
	\includegraphics[width=0.105\textwidth, trim=90 30 50 50, clip]{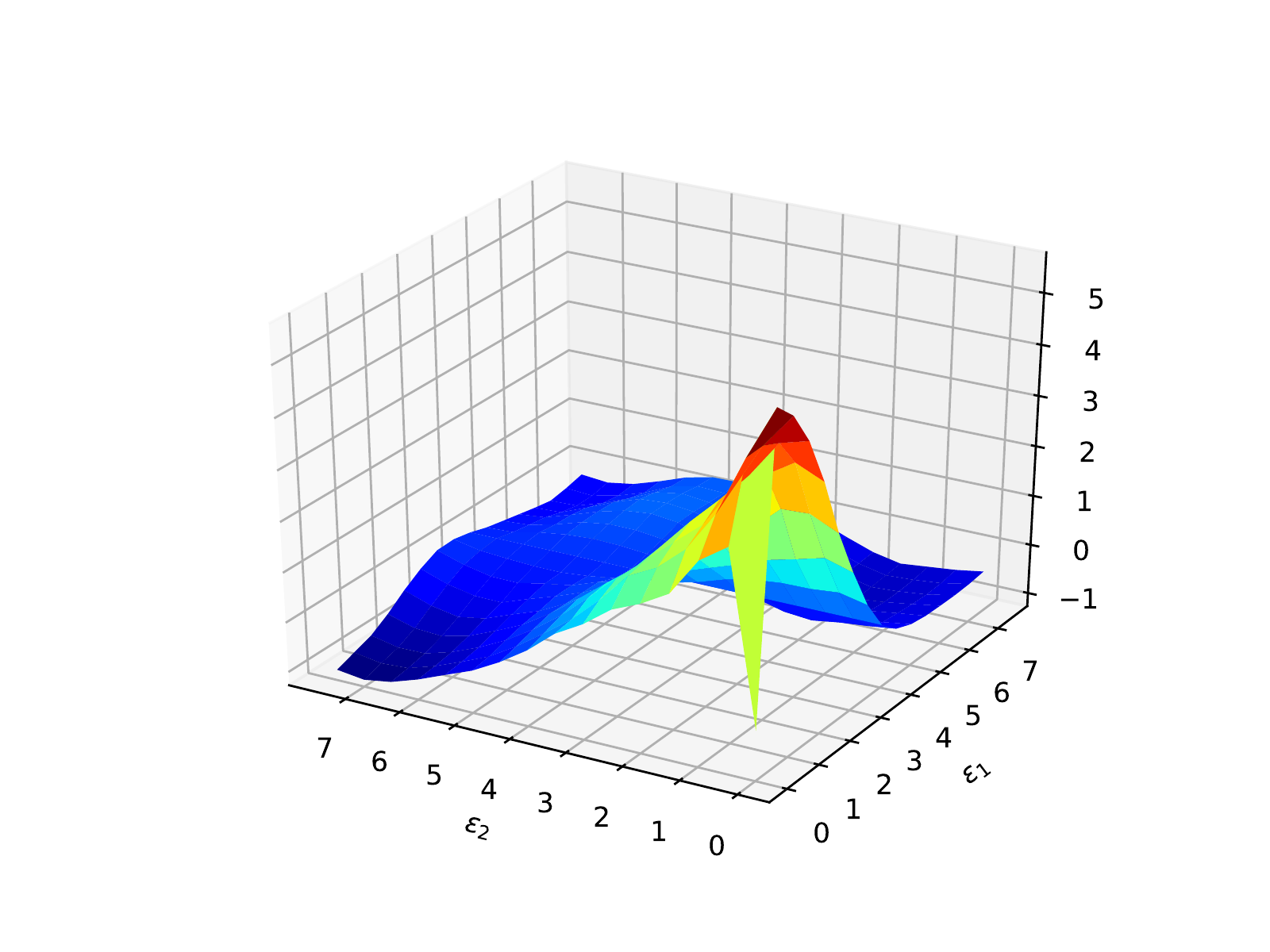} %
	\includegraphics[width=0.105\textwidth, trim=90 30 50 50, clip]{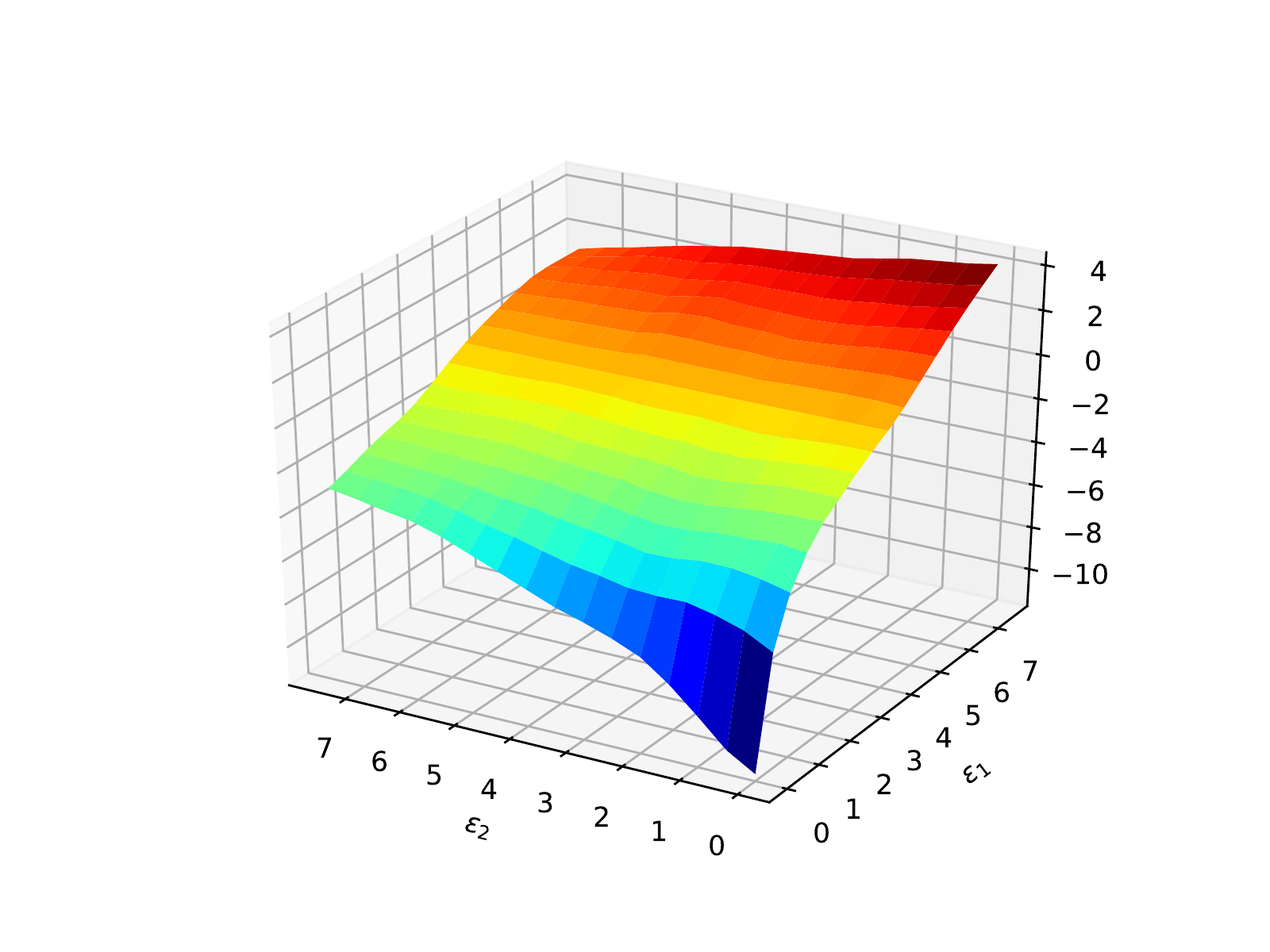}
	\caption{The landscape (bottom row) of the loss around the input examples (top row). The $x,y$-axis of the loss landscape plots are $\epsilon_1$ and $\epsilon_2$, which are the sizes of perturbations added to two adversarial directions $\gb$ and $\gb^{\perp}$ respectively: $\xa = \xb + \epsilon_1 \gb + \epsilon_2 \gb^{\perp}$, where $\gb$ is the adversarial direction (sign of the input gradients) and $\gb^{\perp}$ is the adversarial direction found from the surrogate models. The $z$-axis of the loss landscape is the classification loss. The use of overparameterized deep networks on medical images causes the loss landscapes around medical images extremely sharp, compared to that of natural images.}
	\label{fig:loss_landscape}
\end{figure*}

\textbf{DNN Model Viewpoint.} We next show that the higher vulnerability of medical DNN models is largely caused by the use of overparameterized deep networks for simple medical image analysis tasks. The third row in Figure \ref{fig:saliency} illustrates the representations learned at an intermediate layer of ResNet-50, i.e., the averaged `res3a\_relu' layer output over all channels.
Surprisingly, we find that the deep representations of medical images are rather simple, compared to the complex shapes learned from natural images.
This indicates that, on medical images, the DNN model is learning simple patterns (possibly those are only related to the lesions) out of a large attention area. However, learning simple patterns does not require complex deep networks. This motivates us to investigate whether the high vulnerability is caused by the use of overparameterized networks, by exploring the loss landscape around individual input samples. Following previous works for natural adversarial images~\cite{tramer2017space}, we construct two adversarial directions $\gb$ and $\gb^{\perp}$, where $\gb$ and $\gb^{\perp}$ are the input gradients extracted from the DNN classifiers and a set of separately trained surrogate models respectively. We then craft adversarial examples following $\xa = \xb + \epsilon_1 \gb + \epsilon_2 \gb^{\perp}$. More specifically, we gradually increase $\epsilon_1$ and $\epsilon_2$ from 0 to 8.0/255, and visualize the classification loss for each combination of $\epsilon_1$ and $\epsilon_2$ in Figure \ref{fig:loss_landscape}. We observed that the loss landscapes around medical images are extremely sharp, compared to the flat landscapes around natural images. A direct consequence of sharp loss is high vulnerability to adversarial attacks, because small perturbations of an input sample are likely to cause a drastic increase in loss. A sharp loss is usually caused by the use of an over complex network on a simple classification task \cite{madry2018towards}.

In summary, we have found that medical DNN models can be more vulnerable to adversarial attacks compared to natural image DNN models, and we argue this may be due to 2 reasons: 
1) the complex biological textures of medical images may lead to more vulnerable regions; and most importantly, and 2) state-of-the-art deep networks designed for large-scale natural image processing can be overparameterized for medical imaging tasks and result in high vulnerability to adversarial attacks.

\subsection{Discussion}
In deep learning based medical image analysis, it is a common practice to use state-of-the-art DNNs that were originally designed for complex large-scale natural image processing. However, these networks may be overparameterized for many of the medical imaging tasks. We would like to highlight to researchers in the field that, while these networks bring better prediction performance, they are more vulnerable to adversarial attacks. In conjunction with these DNNs, regularizations or training strategies that can smooth out the loss around input samples may be necessary for robust defenses against such attacks.

\section{Understanding the Detection of Medical Image Attacks}
In this section, we conduct various adversarial detection experiments using two state-of-the-art detection methods, \textit{i.e.}, KD \cite{feinman2017detecting} and LID \cite{ma2018characterizing}. In addition, we also investigate the use of deep features (denoted by ``DFeat") or quantized deep features (denoted by ``QFeat") \cite{lu2017safetynet} for adversarial detection. The detection experiments are conducted on our three 2-class datasets. 

\subsection{Detection Settings}
\begin{figure}
    \centering
    \includegraphics[width=\linewidth]{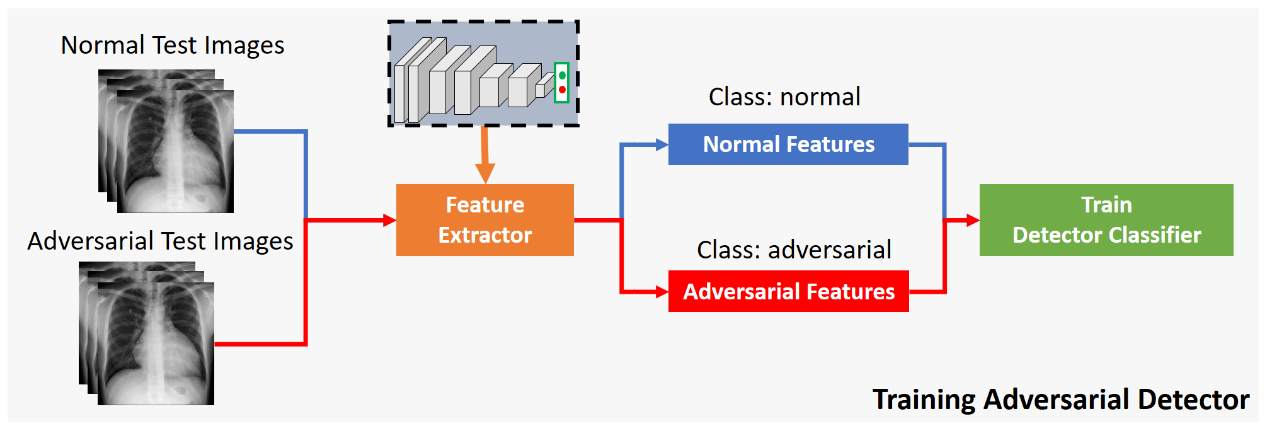}
    \caption{The pipeline of training an adversarial detector.}
    \label{fig:detector_pipeline}
\end{figure}

The DNN models used here are the same as those used in the above attack experiments (see Section \ref{sec:attack}). The detection pipeline is illustrated in Figure \ref{fig:detector_pipeline}. Based on the pretrained DNN models, we apply the four attacking methods (\textit{FGSM, BIM, PGD and CW}) to generate adversarial examples for the correctly classified images from both the \emph{AdvTrain} and \emph{AdvTest} subsets. We then extract the features used for detection, which include the deep features at the second-last dense layer of the network (``DFeat"/``QFeat"), the KD (kernel density estimated from the second-last layer deep features) features, and the LID (local intrinsic dimensionality estimated from the output at each layer of the network) features. All the parameters for KD/LID estimation are set as per their original papers. 
All detection features are extracted in mini-batches of size 100. The detection features are then normalized to [0,1]. The detectors are trained on the detection features of the \emph{AdvTrain} subset, and tested on the \emph{AdvTest} subset.
As suggested by \cite{feinman2017detecting,ma2018characterizing}, we use a logistic regression classifier as the detector for KD and LID, the random forests classifier as the detector for the deep features, and the SVM classifier for quantized deep features. AUC (Area Under Curve) score is adopted as the metric for detection performance. 

\subsection{Detection Results}
We report the detection AUC scores of the 4 types of detectors against the 4 types of attacking methods (white-box) across the three datasets in Table \ref{table:detection}.
State-of-the-art detectors demonstrate very robust performance against these attacks. Especially the KD-based detectors, which achieve an AUC of above $99\%$ against all attacks across all three datasets.
However, on natural images, these state-of-the-art detectors often achieve less than $80\%$ detection AUC against some of the tested attacks such as FGSM and BIM \cite{ma2018characterizing,feinman2017detecting}.
This indicates that medical image adversarial examples are much easier to detect compared to natural image adversarial examples.
Quite surprisingly, we find that the deep features (\textit{e.g.} `DFeat') alone can deliver very robust detection performance against all attacks. In particular, deep feature based detectors achieve an AUC score above $98\%$ across all the testing scenarios. On the other hand, the detectors trained on quantized deep features (\textit{e.g.} `QFeat') also achieve good detection performance. This indicates that the deep features of adversarial examples (adversarial features) may be fundamentally different from that of normal examples (normal features).

\subsection{Detection Transferability}
We further test if the `QFeat' detectors can still have good performance when trained on one attack (source), then applied to detect the other 3 attacks (targets). In this transferability test, we train detectors on `QFeat' of adversarial examples crafted by the source attacks on both \emph{AdvTrain} and \emph{AdvTest} subsets, then apply the trained detectors to detect adversarial examples crafted by other attacks also on both \emph{AdvTrain} and \emph{AdvTest}. As shown in Table \ref{table:detection2}, the detectors trained on either weak attack FGSM or strong attack PGD all transfer perfectly against other attacks. This again confirms that medical image adversarial examples can be easily detected. The 100\% detection AUCs suggests that there are indeed some fundamental differences between adversarial examples and normal examples.

\begin{table}[t]
\renewcommand{\arraystretch}{1.1}
\caption{Detecting white-box attacks: the AUC score (\%) of various detectors against the 4 types of attacks crafted on the three datasets. The best results are highlighted in \textbf{bold}.}
\label{table:detection}
\centering
\small{\resizebox{\linewidth}{!}{
    \begin{tabular}{c|c|cccc}
    \toprule
    Dataset & Detector & FGSM  & BIM   & PGD   & CW \\
    \midrule
    \multirow{4}[2]{*}{Fundoscopy} & KD    & \textbf{100.00} & \textbf{100.00} & \textbf{100.00} & \textbf{100.00} \\
          & LID   & 94.20  & 99.63  & 99.52  & 99.20  \\
          & DFeat & 99.97  & \textbf{100.00} & \textbf{100.00} & 99.99  \\
          & QFeat & 98.87  & 99.82  & 99.91  & 99.95  \\
    \midrule
    \multirow{4}[2]{*}{Chest X-Ray} & KD    & 99.29  & \textbf{100.00} & \textbf{100.00} & \textbf{100.00} \\
          & LID   & 78.40  & 96.92  & 95.20  & 96.74  \\
          & DFeat & \textbf{99.97} & \textbf{100.00} & \textbf{100.00} & \textbf{100.00} \\
          & QFeat & 87.63  & 96.35  & 92.07  & 99.16  \\
    \midrule
    \multirow{4}[2]{*}{Dermoscopy} & KD    & \textbf{100.00} & \textbf{100.00} & \textbf{100.00} & \textbf{100.00} \\
          & LID   & 64.83  & 95.37  & 92.72  & 95.90  \\
          & DFeat & 98.65  & 99.77  & 99.48  & 99.78  \\
          & QFeat & 86.53  & 89.27  & 95.45  & 93.92  \\
    \bottomrule
    \end{tabular}}}
\end{table}

\begin{table}[!t]
\renewcommand{\arraystretch}{1.1}
\caption{The detection transferability of the `DFeat' detector: the AUC score (\%) of the two detectors trained on source attacks FGSM and PGD then applied to detect other 3 attacks. The best results are highlighted in \textbf{bold}.}
\label{table:detection2}
\centering
\small{\resizebox{\linewidth}{!}{
    \begin{tabular}{c|c|cccc}
    \toprule
    Dataset & Source & FGSM  & BIM  & PGD   & CW \\
    \midrule
    \multirow{2}[2]{*}{Fundoscopy} & FGSM  & -- & \textbf{100.00} & \textbf{100.00} & \textbf{100.00} \\
         & PGD & \textbf{100.00}  & \textbf{100.00} &  -- & \textbf{100.00}  \\
         \midrule
   \multirow{2}[2]{*}{Chest X-Ray} & FGSM    &  -- & \textbf{100.00} & \textbf{100.00} & \textbf{100.00} \\
        & PGD & \textbf{100.00}  & \textbf{100.00} &  -- & \textbf{100.00}  \\
         \midrule
    \multirow{2}[2]{*}{Dermoscopy} & FGSM    &  -- & \textbf{100.00} & \textbf{100.00} & \textbf{100.00} \\
        & PGD & \textbf{100.00}  & \textbf{100.00} &  -- & \textbf{100.00}  \\
    \bottomrule
    \end{tabular}}}
\end{table}

\subsection{Why are Adversarial Attacks on Medical Images Easy to Detect?}

\begin{figure}
	\centering
	\resizebox{\linewidth}{!}{\begin{tabular}{ccccc}
    \begin{sideways}\hspace{3mm}\small \textsf{Fundoscopy}\end{sideways} &
    \includegraphics[width=0.3\linewidth]{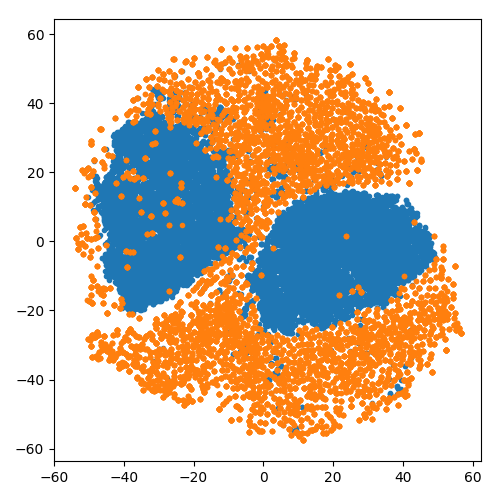} &
    \includegraphics[width=0.3\linewidth]{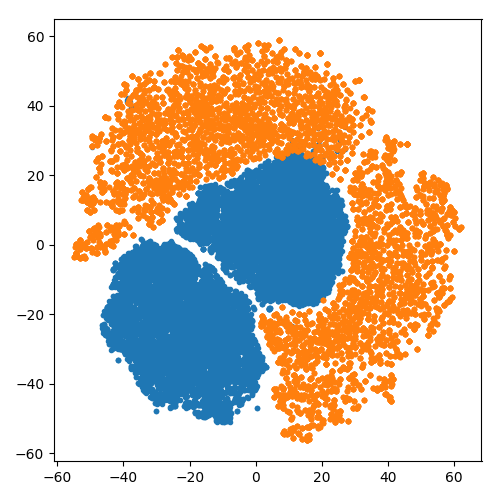} &
    \includegraphics[width=0.3\linewidth]{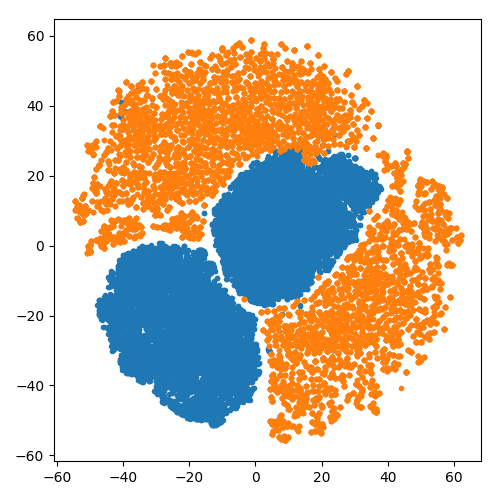} & 
    \includegraphics[width=0.3\linewidth]{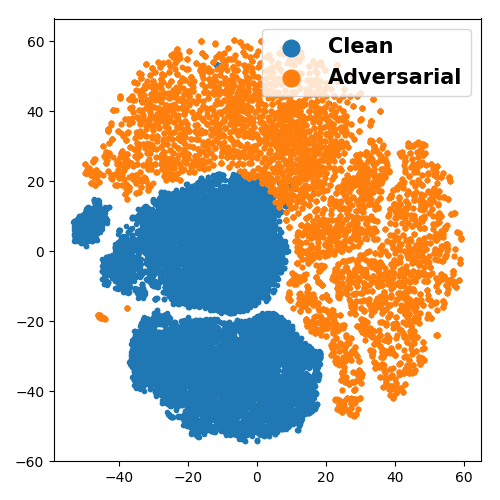} \\
    \begin{sideways}\hspace{3mm}\small \textsf{Chest X-Ray}\end{sideways} &
    \includegraphics[width=0.3\linewidth]{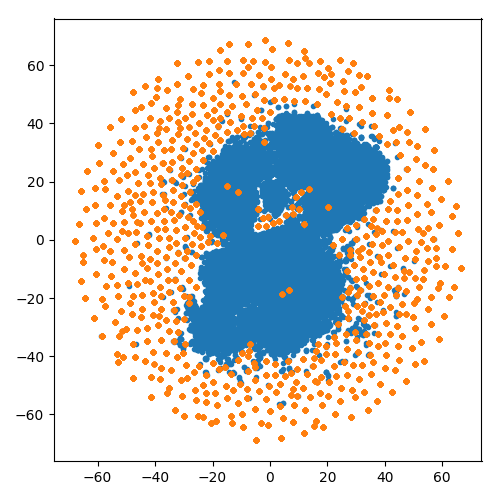} &
    \includegraphics[width=0.3\linewidth]{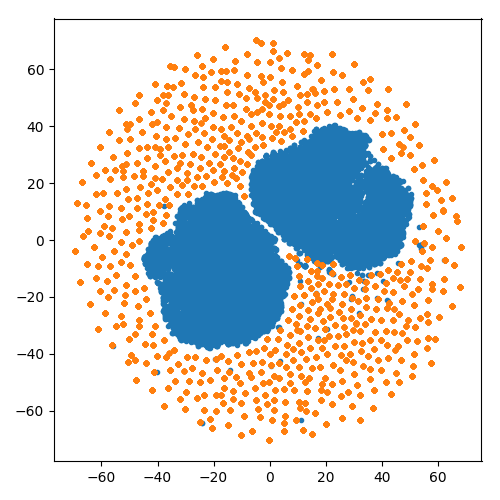} &
    \includegraphics[width=0.3\linewidth]{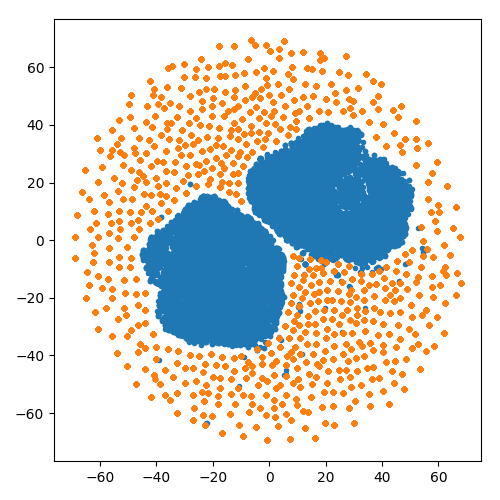} & 
    \includegraphics[width=0.3\linewidth]{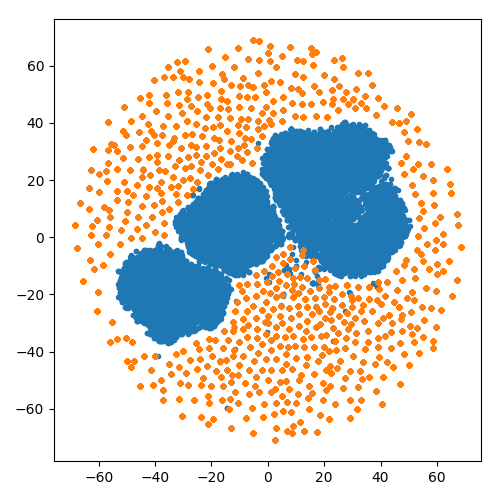} \\
    \begin{sideways}\hspace{3mm}\small \textsf{Dermoscopy}\end{sideways} &
    \includegraphics[width=0.3\linewidth]{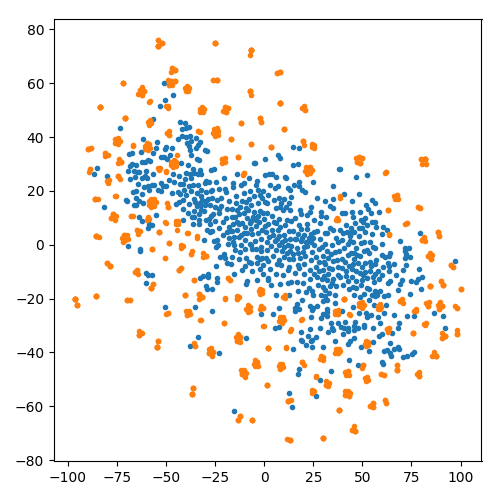} &
    \includegraphics[width=0.3\linewidth]{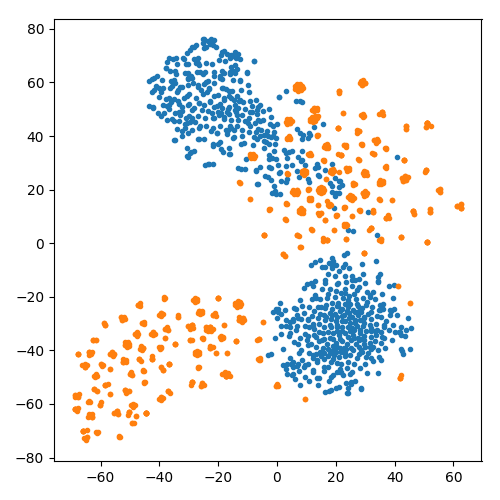} &
    \includegraphics[width=0.3\linewidth]{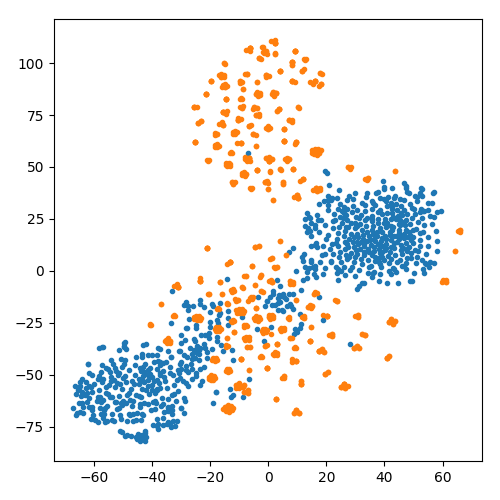} & 
    \includegraphics[width=0.3\linewidth]{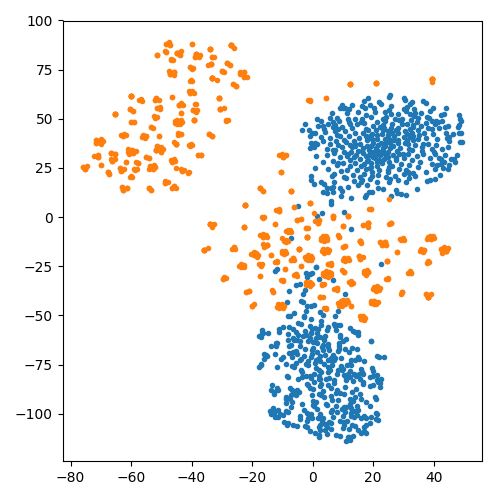} \\
          & \small \textsf{Clean vs FGSM} & \small \textsf{Clean vs BIM} & \small \textsf{Clean vs PGD} & \small \textsf{Clean vs CW} \\
    \end{tabular}}%
	\caption{Visualization of t-SNE 2D embeddings of adversarial and normal features, extracted from the second last dense layer of the DNN models. Each row is a dataset, each column is an attack, and blue/orange indicates clean and adversarial examples respectively.}
	\label{fig:feat_pca}
\end{figure}

To better illustrate the difference between adversarial and normal features, we visualize the 2D embeddings of the deep features using t-SNE \cite{maaten2008visualizing}. We observe in Figure \ref{fig:feat_pca} that adversarial features are almost linearly separable (after some non-linear transformations) from normal features. This is quite different from natural images, where deep features of adversarial examples are quite similar to that of normal examples, and deep feature based detectors can only provide limited robustness \cite{feinman2017detecting,ma2018characterizing}.

\begin{figure*}[!t]
	\centering
	\includegraphics[width=0.8\textwidth]{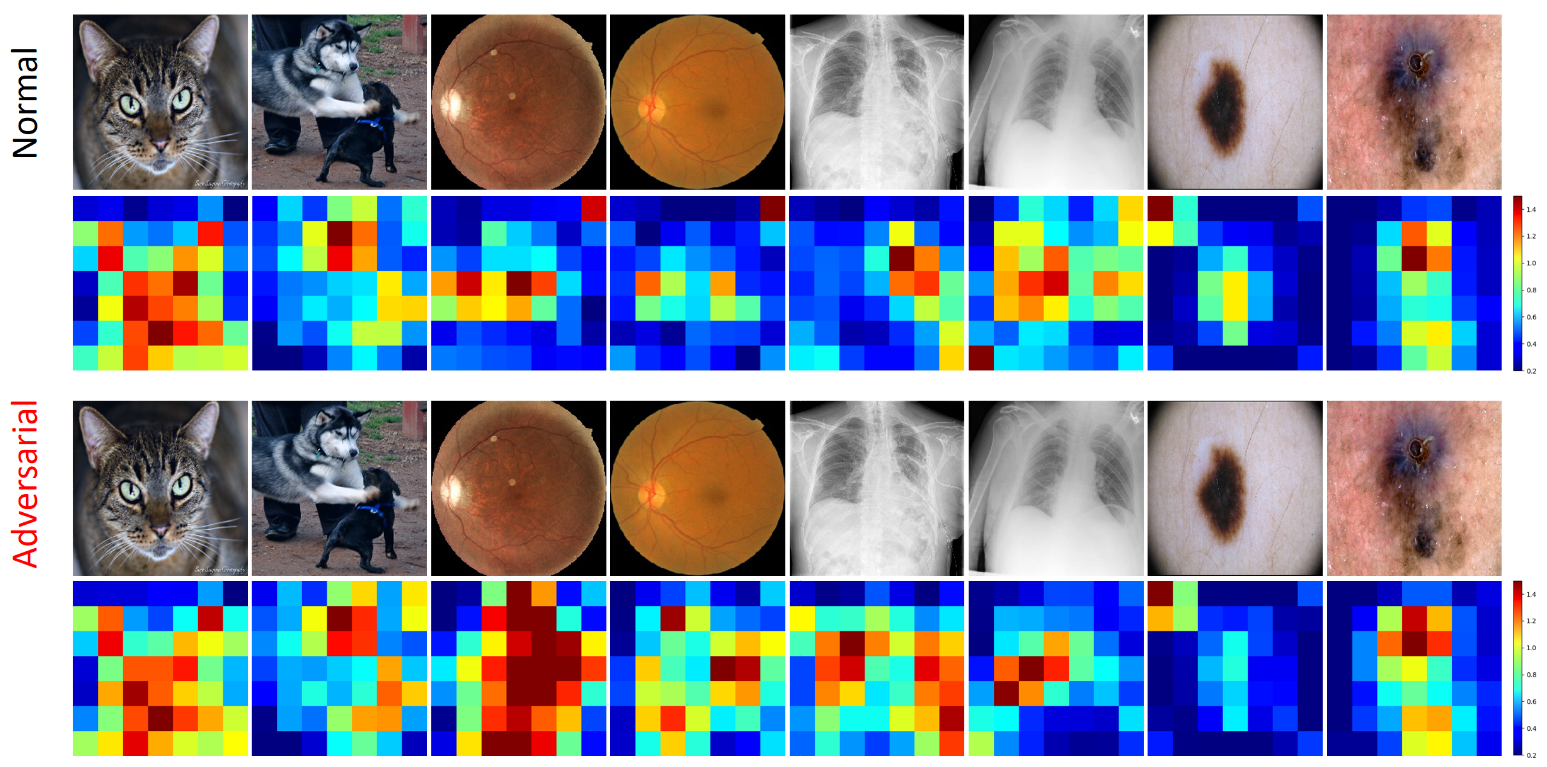}
	\caption{The deep representations on normal images (first row) versus adversarial images (third row) learned by the ResNet-50 models at the `res5b\_relu' layer (averaged over channels).}
	\label{fig:advs_feature}
\end{figure*}

Similar to Figure \ref{fig:saliency}, we visualize the deep representation of normal and adversarial examples in Figure \ref{fig:advs_feature}.
Here, we focus on features learned at a deeper layer (eg. the `res5b\_relu' layer of ResNet-50), as we are more interested in the cumulative effect of adversarial perturbations.
We find that there are clear differences between adversarial and normal representations, especially for medical images. Compared to natural images, adversarial perturbations tend to cause more significant distortions on medical images in the deep feature space.
Considering the difference in deep representations between natural images and medical images (Figure \ref{fig:saliency}), this will lead to effects that are fundamentally different for natural versus medical images.
As the deep representations of natural images activate a large area of the representation map, the adversarial representations that are slightly distorted by adversarial perturbations are not significant enough to be different from the normal representations. However, the deep representations of medical images are very simple and often cover a small region of the representation map. We believe this makes small representation distortions stand out as outliers.

\begin{figure*}[!t]
	\centering
	\includegraphics[width=0.8\textwidth]{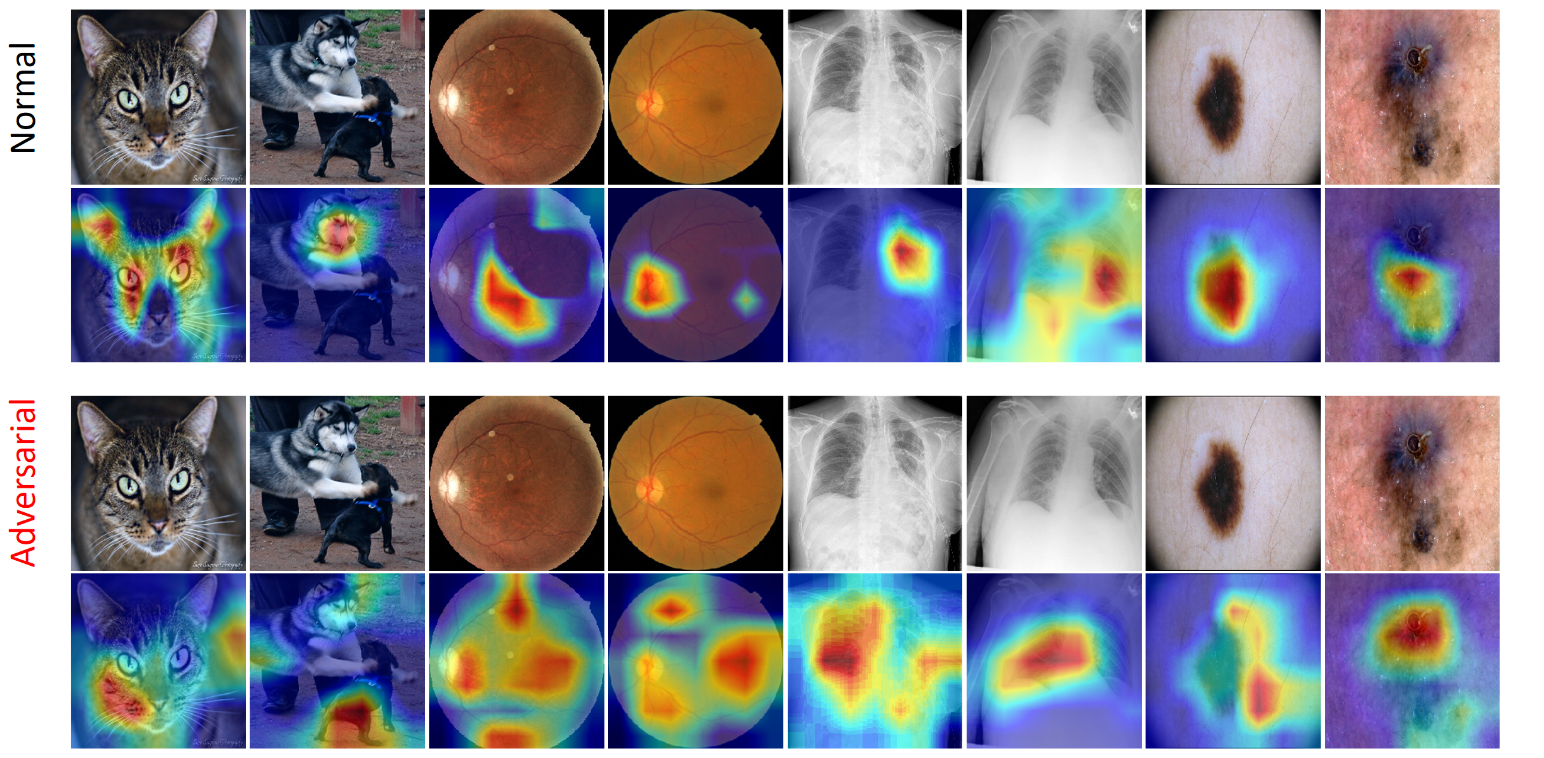}
	\caption{The attention maps of the network on normal images (first row) versus adversarial images (third row). The attention maps are computed by the Grad-CAM technique \cite{selvaraju2017grad}.}
	\label{fig:cam}
\end{figure*}

To further understand why tiny changes in deep features can make a fundamental difference, we show the attention maps of both normal and adversarial examples in Figure \ref{fig:cam}. 
We exploit the Gradient-weighted Class Activation Mapping (Grad-CAM) technique \cite{selvaraju2017grad} to find the critical regions in the input image that mostly activate the the network output.
Grad-CAM uses the gradients of a target class, flowing into the final convolutional layer to produce a coarse localization map highlighting the important regions in the image for predicting the class. As demonstrated in Figure \ref{fig:cam}, the attentions of the DNN models are heavily disrupted by adversarial perturbations. On natural images, the attentions are only shifted to less important regions which are still related to the target class. For example, in the `cat' example, the attention is shifted from the ear to the face of the cat. However, on medical images, the attentions are shifted from the lesion to regions that are completely irrelevant to the diagnosis of the lesion. This explains why small perturbations in medical images can lead to deep features that are fundamentally different and easily separable from the normal features.

\subsection{Discussion}
According to our above analysis, medical image adversarial examples generated using attacking methods developed from natural images are not really ``adversarial" from the pathological sense. Careful consideration should be made if using these adversarial examples to evaluate the performance of medical image DNN models. Our study also sheds some light on the future development of more effective attacks on medical images. Pathological image regions might be exploited to craft attacks that produce more misleading adversarial features that are indistinguishable from normal features. Such attacks might have a higher chance to fool both the DNN models and the detectors.

\section{Discussion and Conclusion}\label{sec:conclusion}
\subsection{Discussion}

Although existing attacks can easily fool deep neural networks (DNNs) used for medical image analysis, the perturbations are small and imperceptible to human observers, thus posing very limited impact on the diagnosis results when medical experts are involved.  Whether physical world medical image examples can be crafted to fool both deep learning medical systems and medical experts is still not clear. While it has been demonstrated possible on natural images \cite{kurakin2017adversarial}, traffic signs \cite{evtimov2017robust} or object detectors \cite{liu2019perceptual}, the crafted adversarial stickers or patches are obviously malicious to humans. We believe more subtle and stealthy perturbations will be required for physical-world medical image adversarial examples. 

On the defense side, effective defense techniques against medical image adversarial examples are imperative. While existing defense methods developed on natural images such as adversarial training \cite{madry2018towards,wang2019convergence,Wang2020Improving,yu2019towards} and regularization methods \cite{ross2018improving,zhang2019interpreting} may also apply for medical image adversarial examples, more effective defenses might be developed by also addressing the overparameterization of DNNs used in deep learning medical systems.

\subsection{Conclusion}

In this paper, we have investigated the problem of adversarial attacks on deep learning based medical image analysis. A series of experiments with 4 types of attack and detection methods were conducted on three benchmark medical image datasets. We found that adversarial attacks on medical images are much easier to craft due to the specific characteristics of medical image data and DNN models. More surprisingly, we found that medical adversarial examples are also much easier to detect, and that simple deep feature based detectors can achieve over 98\% detection AUC against all tested attacks across the three datasets and detectors trained on one attack transfer well to detect other unforeseen attacks. This is because adversarial attacks tend to attack a widespread area outside the pathological regions, which results in deep features that are fundamentally different and easily separable from normal features. 

Our findings in this paper can help understand why a deep learning medical system makes a wrong decision or diagnosis in the presence of adversarial examples, and more importantly, the difficulties in generating and detecting such attacks on medical images compared to that on natural images. This can further motive more practical and effective defense approaches to improve the adversarial robustness of medical systems.
We also believe these findings may be a useful basis to approach the design of more explainable and secure medical deep learning systems.

\section*{Acknowledgement}

This work was supported by National Natural Science Foundation of China (NSFC) under Grant 61972012 and JST, ACT-X Grant Number JPMJAX190D, Japan and  Zhejiang Provincial Natural Science Foundation of China (LZ19F010001).

\bibliography{arxiv}
\bibliographystyle{icml2019}

\end{document}